\documentclass[12pt]{article}

\usepackage{newtxtext,newtxmath}

\usepackage{graphicx}

\usepackage[letterpaper,margin=1in]{geometry}

\renewenvironment{abstract}
	{\quotation}
	{\endquotation}

\date{}

\makeatletter
\renewcommand{\fnum@figure}{\textbf{Figure \thefigure}}
\renewcommand{\fnum@table}{\textbf{Table \thetable}}
\makeatother

\usepackage{scicite}

\usepackage{url}

\def\scititle{
	Expressive Robotic Pianist: Mastering Complex Piano Repertoire with Graph-Mimic and Musical Dynamics
}
\title{\bfseries \boldmath \scititle}

\author{
	Yanhong Liang$^{1}$,
	Xianwei Liu$^{2}$,
	Chaojie Fu$^{1}$,
	Shaowen Cheng$^{2}$,
	Yanyan Yuan$^{1}$,\and
	Chengwei Zhuo$^{1}$,
	Xi Chen$^{3}$,
	Yongbin Jin$^{2\ast}$,
	Wei Yang$^{1\ast}$,
	Hongtao Wang$^{1\ast}$\and
	\small$^{1}$the Center for X-Mechanics, Zhejiang University, Hangzhou \& 310027, China.\and
	\small$^{2}$ZJU-Hangzhou Global Scientific and Technological Innovation Center, Hangzhou \& 311200, China.\and
	\small$^{3}$Department of Public Physical and Art Education, Zhejiang University, Hangzhou \& 310058, China.\and
	\small$^\ast$Corresponding author. Email: yongbinjin@zju.edu.cn\and
}

\begin{document} 

\maketitle

\begin{abstract} \bfseries \boldmath

Enabling robots to perform musical instruments with human-level expressivity represents a frontier 
in bridging the gap between mechanical precision and artistic interpretation. Despite advances in 
robotic dexterity, replicating the fluid finger transitions and nuanced dynamic control characteristic 
of human pianists remains a significant challenge.
Through a reinforcement learning–based control framework, we demonstrate that a dexterous robotic 
hand can achieve high-fidelity performance across a diverse piano repertoire.Central to our approach is a 
graph-based optimization strategy that guids the robot to generate natural pre-press and key-press fingering 
strategies that closely resemble human movement patterns. To achieve expressive sound production, the control 
system is coupled with a physics-inspired acoustic model that modulates keypress velocity to accurately reproduce 
the dynamic variations specified in musical scores.
Quantitative evaluations demonstrate that our expressive control model significantly outperforms baseline 
methods in both finger morphology similarity and dynamic velocity accuracy. In a perceptual test involving 
participants from diverse listener groups, performances generated by our system are significantly preferred 
over baseline robotic performances and are indistinguishable from human performances for non-professional 
audiences. Furthermore, extensive experiments across multiple musical styles confirm that our method maintains 
high note-level accuracy while achieving expressive performance. 
In conclusion, our approach provides a robust pathway for robotic systems to move beyond mere mechanical accuracy, 
elevating robotic musicianship to a level of expressive performance comparable to human pianists.


\end{abstract}

\noindent
\textbf{One-Sentence Summary}: A dexterous robotic pianist delivers human-level artistic performance 
with natural finger coordination and expressive dynamics.

\newpage
\section*{Introduction}

\noindent
Replicating piano performance is among the most challenging tests of robotic dexterity. 
The piano conveys the full spectrum of a musical composition, including melody, harmony, 
bass, form, and intricate counterpoint, through dexterous hand movements\cite{hofmann1976piano}. 
Mastery requires expressive control of dynamics, articulation, and timing, and even human 
pianists need years of intensive practice to acquire these skills and overcome performance 
ceilings\cite{furuya2025surmounting}. Achieving comparable artistry in robotic hands is 
exceptionally difficult due to mechanical constraints and the lack of musically aware 
control strategies. In this work, we demonstrate a robotic hand capable of performing piano 
literature at the Grade 7 standard, meeting the repertoire's rigorous demands for fingering 
agility, articulatory touch, and dynamic expression. This work represents an incremental move 
toward bridging the gap between robotic dexterity and human-level motor expressiveness.

Expressive piano performance is defined by the synthesis of three competencies: 
fluid fingering transitions during pre-press phase, articulation-specific touch during key-press phase, 
and precise dynamic sound control. 
Fluid fingering transitions, grounded in pedagogical practice, rely on sequencing that 
prioritizes movement efficiency and gestural fluency, a challenge acutely evident 
in passages with wide spans or rapid scales. Finger-key touch shapes articulation, 
where vertical strikes produce staccato and shallow, gliding 
contact enables legato passages. Numerous methodology books and technical exercises 
have been developed to systematically train these skills\cite{furuya2011hand,neuhaus2008art,sandor1981piano,chang2016fundamentals}.
Control of keypress velocity governs musical dynamics, allowing performers to realize 
crescendos, decrescendos, and subtle contrasts, thereby conveying emotional intent and 
musical structure. In robotic execution, this requires both mechanical capability and a 
control strategy that jointly regulate fingertip speed to achieve intended 
loudness\cite{gieseking2013piano,slenczynska1968music}. Together, these elements define the 
pedagogical foundation of expressive playing and set the technical requirements for a 
robotic hand to replicate human-like artistry.

\subsubsection*{Related work}
Prior robotic systems have enabled automated piano performance on relatively simple pieces. 
Early approaches relied on pre-programmed strategies. For example, WABOT-2 employed a 
14-degree-of-freedom(DoF) robotic hand to perform basic piano tasks\cite{sugano1987wabot}, 
while TeoTronico, equipped with 53 mechanical fingers, was capable of covering the entire 
keyboard\cite{spectrum2017music}. Some designs used five-fingered 
hands to autonomously play electronic pianos from user-defined sheet 
music\cite{lin2010electronic,li2013controller,li2014intelligent}, some focused on force 
modulation through linear motor control to shape intonation\cite{li2015force}, and some applied 
trajectory optimization to enable the Shadow Hand\cite{shadowrobot2025} to perform on a 
physical piano\cite{scholz2019playing}. More recent systems introduced skeletal hands with 
style-specific adaptability\cite{hughes2018anthropomorphic} or biomimetic rigid-soft 
finger structures\cite{zhang2025biomimetic}. Despite these advances, most approaches 
emphasize striking the correct keys at the correct times, while maintaining fixed, 
uniform postures and articulation. As emphasized in \textit{Music at Your Fingertips}\cite{slenczynska1968music}, 
"A true pianist's touch is ever-changing, capable of generating endless variations in 
tone color modulated by both hands." Pre-programmed strategies typically assign a single 
configuration per finger-key pair, ignoring the rich repertoire of feasible postures 
offered by multi-DoF hands and suppressing the natural variability that underlies expressive 
performance. Consequently, critical aspects of pianism, including fingering transition strategies, 
finger-key touch techniques, and nuanced dynamic control, cannot be fully captured by 
rigid pre-programming alone.

Reinforcement learning (RL) offers a promising alternative, allowing robotic agents 
to autonomously explore diverse control strategies. Early efforts incorporated tactile 
feedback\cite{xu2022towards},enabling the Allegro Hand\cite{allegrohand2025} to 
perform simple note sequences in simulation. Robopianist\cite{robopianist2023} 
exemplifies this approach, training a pair of Shadow Hands in MuJoCo\cite{todorov2012mujoco} 
to play multiple pieces with correct timing and note durations through a carefully 
designed reward structure.These studies demonstrate RL's potential for 
robotic piano performance and have inspired numerous subsequent improvements
\cite{qian2024pianomime,zhao2024rpm,yang2024learning,huang2025pandora,abe2024latent,li2024humanoid}.

Despite the progress, practical deployment remains constrained by three fundamental challenges.
Pre-press fingering coordination is primarily a kinematic problem that requires smooth planning of 
finger replacement to preserve temporal continuity and musical phrasing. 
Prior work \cite{qian2024pianomime} incorporates human motion references based on fingertip positions 
to guide this process, improving key selection. Moreover, key-press behavior is a contact-dynamic problem, 
in which fingertip orientation and equivalent contact stiffness shape articulation. 
The unique contact–detach cycles in piano playing require motion representations to 
account for both fingertip trajectories and intermediate joint configurations, which 
are essential for optimizing pre-press coordination and modulating key-press equivalent 
stiffness.
Loudness control further requires precise 
regulation of contact velocity to achieve intended intensity and expressive nuance. 
Many existing RL formulations underrepresent musical dynamics, despite its critical role in 
achieving expressive tone production \cite{scimeca2020gaussian}, and do not consider how finger 
motion and contact mechanics influence sound production, leading to suboptimal convergence and 
limiting the emergence of human-like postures and stable articulation.

\subsubsection*{Contributions}
\begin{figure}
    \centering
    \includegraphics[width=1.0\textwidth]{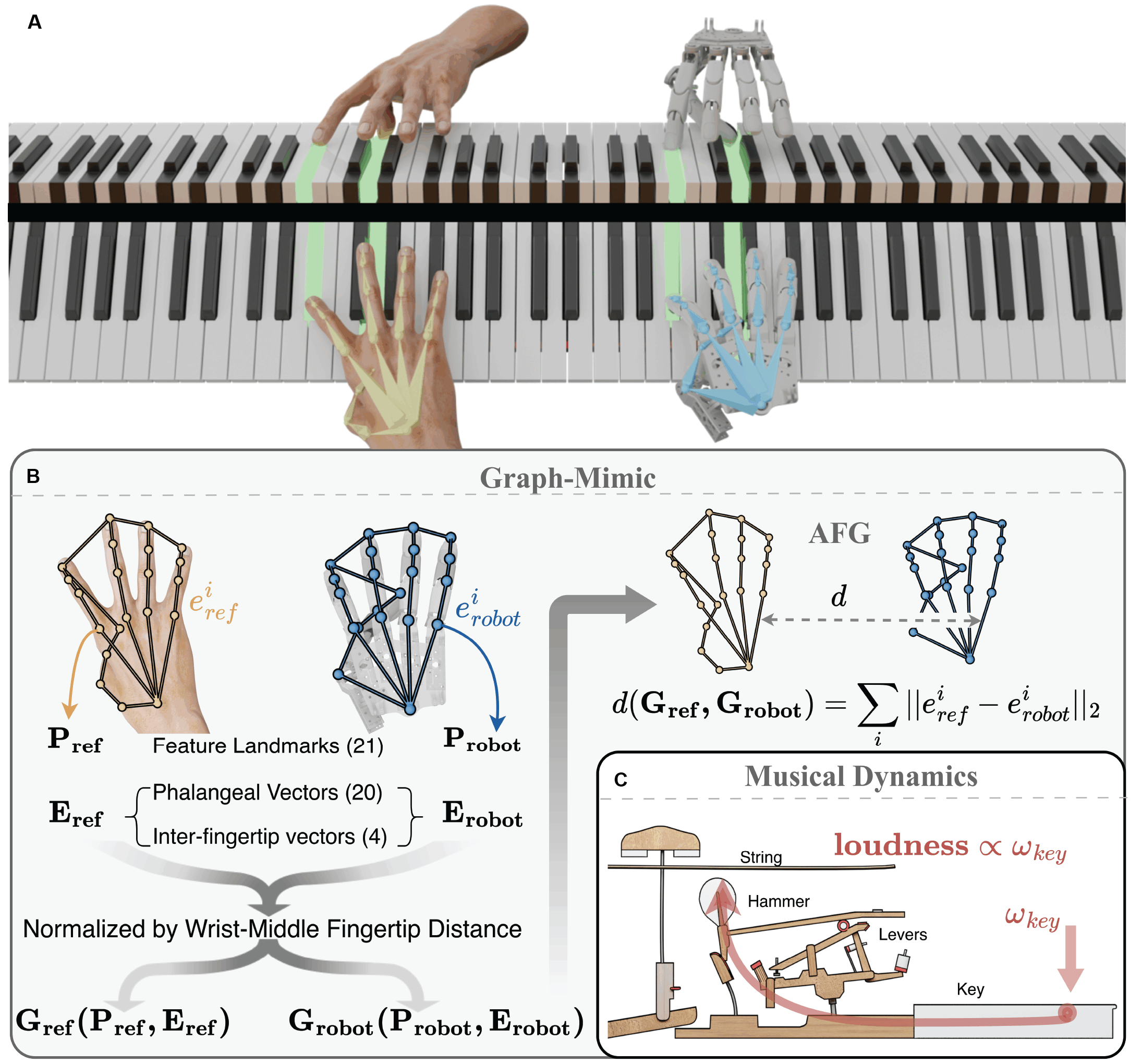}
	\caption{\textbf{Overview of the robotic pianist system and core models.}
    \textbf{(A)} An 18DOF robotic hand reproduces human pianistic techniques to perform on a piano.
    \textbf{(B)} Graph-Mimic, guiding the robot's hand posture to imitate that of a human.
    \textbf{(C)} Musical Dynamics model, establishing the relationship between key velocity ($\omega_{key}$) and note loudness.
    }\label{Overallintro}
\end{figure}

To overcome these challenges and bridge the gap between mechanical note execution and artistic expression 
(Fig.~\ref{Overallintro}A), we introduce an integrated framework combining motion 
imitation with acoustic modeling: Graph-Mimic and Musical Dynamics.

First, to quantify and optimize the motion similarity between human and robotic hands 
we introduce Graph-Mimic, a morphology-agnostic 
retargeting framework inspired by prior advances in character control 
\cite{peng2018deepmimic,peng2021amp,tang2024humanmimic,he2025asap,zhang2023simulation} 
and robotic manipulation \cite{handa2020dexpilot, yang2025omniretarget}. Human perceptual studies have shown that 
motion similarity is judged primarily through spatial joint relationships rather than raw 
joint orientations \cite{tang2008emulating,basset2022impact,harada2004quantitative,chen2010learning}. 
Following this principle, Graph-Mimic represents hand motion as normalized spatial 
relationship graphs constructed from shared feature points, eliminating the need for 
strict joint-to-joint correspondence. This representation, defined as the Action Frame Graph (AFG) 
and depicted in Fig.~\ref{Overallintro}B, establishes a mathematically rigorous yet perceptually 
valid foundation for aligning robotic motion with human biomechanical patterns. 
The framework optimizes a perceptually grounded similarity metric through two complementary 
mechanisms within the AFG: Inter-fingertip Vectors that shape hand morphology 
during the pre-press phase to optimize fingertip trajectories, effectively prioritizing 
finger dexterity over wrist repositioning; and Phalangeal Vectors that 
govern contact configuration during the key-press phase, modulating the effective 
stiffness along the actuation axis to shape articulatory characteristics 
(e.g., staccato vs. legato).

Second, as illustrated in Fig.~\ref{Overallintro}C, we introduce a simplified piano acoustics 
model to guide the robot in controlling Musical Dynamics. The sound production in a piano, 
as a typical string instrument, involves the complex vibration of strings struck by hammers. 
Such detailed simulation presents a major challenge for RL algorithm, which rely on large-scale 
sampling. Instead of modeling string vibration directly, we focus on the energy transfer during 
key presses, which significantly simplifies the computation (text S3). 
Studies\cite{furuya2008expertise,fletcher2012physics,wang2024human} 
show that note loudness correlates with key velocity: higher velocity delivers more hammer 
energy, producing a louder tone. This model establishes a quantitative relationship between 
the key angular velocity and the resulting note volume, consistent with findings from physical experiments  
\cite{fletcher2012physics}. Leveraging it, the robot learns 
both when and how fast to press keys, enabling expressive performance while avoiding 
inaudible or excessively forceful strikes.

We evaluated our method on the InReal robotic hand (section \textbf{Dexterous Robotic Hand}) across a 
repertoire of piano pieces varying in difficulty, duration, and style. The system 
demonstrates robust performance by generating musically appropriate fingering and 
nuanced loudness modulation, achieving a level of expressivity comparable to human 
performance (Movie 1). Beyond objective metrics, we conducted a human perceptual evaluation. 
For general listeners, our expressive robotic system demonstrated a level of performance 
comparable to skilled human pianists, with no statistically significant preference found 
between the two groups.

\section*{Results}
\makeatletter
\newcommand{\rom}[1]{\romannumeral #1}
\makeatother


\subsubsection*{Fingering techniques during pre-press phase}

\begin{figure}
\centering
\includegraphics[width=1.0\textwidth]{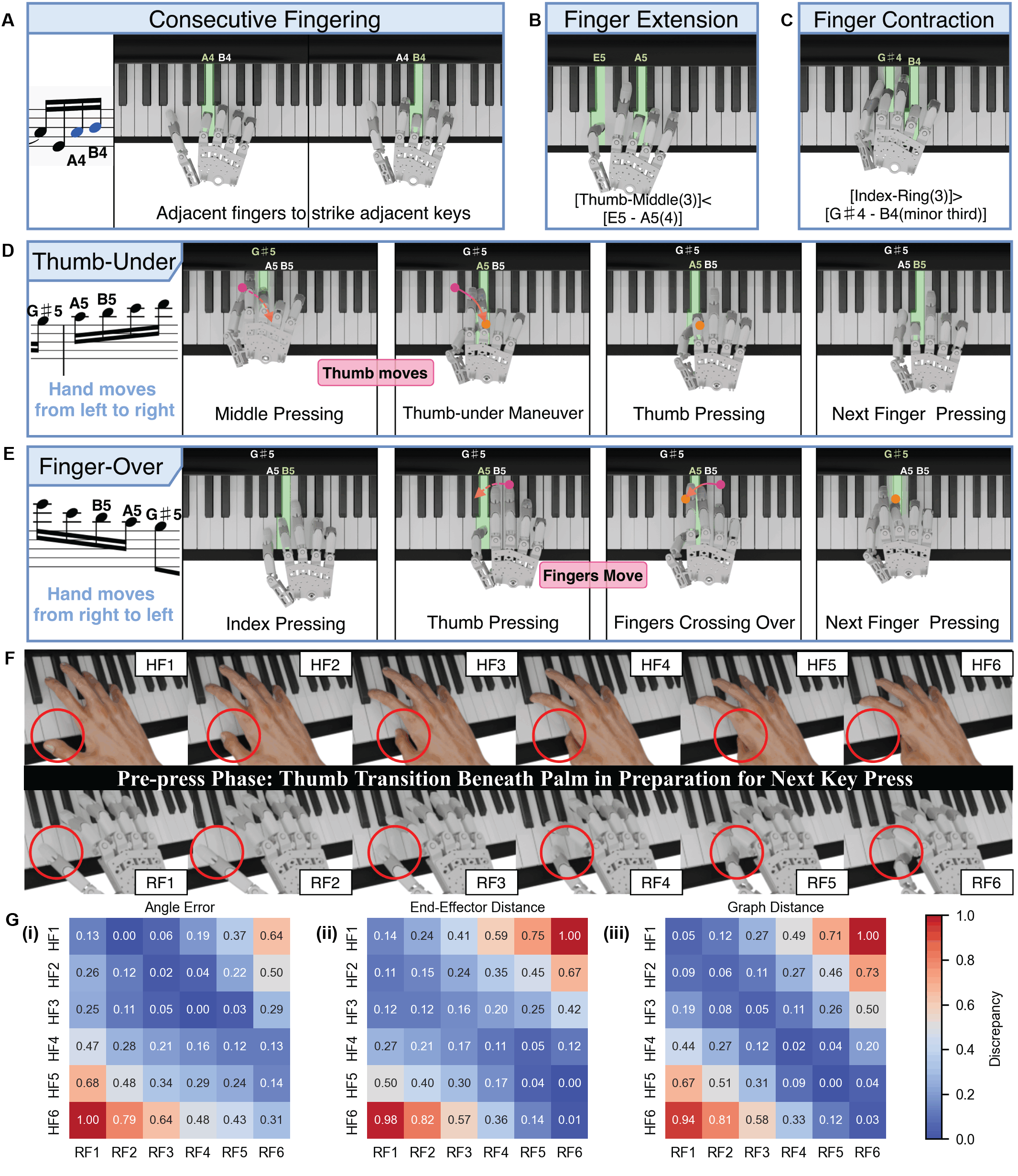}
\caption{\textbf{Graph-Mimic results for pre-press performance.}
\textbf{(A)}–\textbf{(E)} Five essential piano fingering techniques learned by the robotic hand: \textbf{(A)}Consecutive fingering, \textbf{(B)} Finger extension, \textbf{(C)} Finger contraction, \textbf{(D)}Thumb-under, and \textbf{(E)} Finger-over.
\textbf{(F)} Thumb-under motion snapshots from human hand (HF1-HF6) and robotic hand (RF1-RF6).
\textbf{(G)} Normalized discrepancy between human hand and roboitic hand calculated by 
(\rom{1}) joint angle error, (\rom{2}) end-effector distance, and (\rom{3}) graph distance.
}\label{Graphmimic_first}
\end{figure}

We selected \textit{River Flows in You }(Supplementary Movie S1) to highlight fingering 
strategies during key pre-press phase. Consecutive fingering occurs when adjacent fingers press 
neighboring keys in sequence (Fig.~\ref{Graphmimic_first}A, index to middle on A5-B5). 
The hand adjusts its span according to the interval, extending to cover wider 
distances (Fig.~\ref{Graphmimic_first}B, thumb to middle on E5-A5) or contracting when 
more fingers are needed over a narrower range, as depicted in Fig.~\ref{Graphmimic_first}C, 
the index to ring fingers pressing G$\sharp$4-B4 to form a minor third, with an inward 
adjustment to accommodate the black key.

The Thumb-Under Technique(Fig.~\ref{Graphmimic_first}D), used for ascending scales, extends 
the hand's range by tucking the thumb under the fingers(e.g.,thumb reaching A5 after the 
middle finger presses G$\sharp$5, followed by the index finger on B5).
Conversely, the Finger-Over Technique(Fig.~\ref{Graphmimic_first}E), used for descending scales, involves crossing a longer finger over the thumb to reach lower notes 
(e.g., the middle finger crossing over the thumb after pressing A5 to play G$\sharp$5). Since these pre-press transitions are crucial for technical 
fluency and expressive quality, and highly depend on fingertip spatial positions, our proposed Graph Distance metric is essential.
By incorporating inter-fingertip vectors within the AFG (Fig.~\ref{Overallintro}B), the metric provides an effective 
means to guide the robotic hand in mastering these fingering techniques.

To substantiate the advantage of our method, we compared it against two common similarity metrics: Joint Angle Error 
and End-Effector Distance, using the thumb-under motion (Fig.~\ref{Graphmimic_first}F, human movements HF1-6 and 
corresponding robotic executions RF1-6)as a test case. This motion, critical for thumb repositioning, 
heavily relies on fingertip spatial positioning. The results are shown in the similarity matrices(Fig.~\ref{Graphmimic_first}G). 
Both End-Effector Distance and Graph Distance (Fig.~\ref{Graphmimic_first}G (\rom{2}),(\rom{3}))show clear diagonal alignments, 
reflecting accurate spatial imitation. In contrast, the Joint Angle Error(Fig.~\ref{Graphmimic_first}G (\rom{1})) 
fails to capture this correspondence, as angular discrepancies do not align with perceptual motion similarity.

The underlying reason lies in the information captured by each metric. 
End-Effector Distance, which calculates the positional error between 
corresponding fingertips in 3D space, and our Graph Distance, which contains inter-fingertip vectors, 
both inherently encode fingertip spacial relationships. In contrast, Joint Angle Error, computed as the 
difference in angles between corresponding joint pairs, is highly sensitive to hand morphology. A given set of joint angles 
produces drastically different fingertip positions on hands of different sizes and proportions, making this metric a 
poor proxy for functional similarity in the task space.

\subsubsection*{Articulation-specific touch techniques during key-press phase}

Articulation style is crucial for conveying musical character. Staccato and legato are two 
representative styles: staccato evokes a crisp, percussive texture, where each note is released before the next is pressed; 
legato produces a continuous, singing quality, where the subsequent note begins before the previous one is fully released. 
These styles are governed not only by timing but also by finger contact and force modulation. With Graph-Mimic guidance, the 
robotic hand differentiates these techniques: in staccato, illustrated by the spectrogram of \textit{Twinkle Twinkle Little Star} 
(Fig.~\ref{Graphmimic_second}A(\rom{1}), movie S2), the robot favors vertical fingertip contact (Fig.~\ref{Graphmimic_second}A(\rom{2})); 
in legato, shown with \textit{Ode to Joy} (Fig.~\ref{Graphmimic_second}B(\rom{1}), movie S3), it achieves overlapping depressions through stable pad 
contact (Fig.~\ref{Graphmimic_second}B(\rom{2})).

To examine the underlying mechanics, the finger was modeled as a three-link chain with 
metacarpophalangeal (MCP), proximal interphalangeal (PIP), and distal interphalangeal (DIP) joints 
(see detailed parameters in text S1, fig. \ref{fig:S1} and table. \ref{tab:s1_finger_params}). When these joints flex synchronously, the linkage geometry produces 
a systematic increase in equivalent stiffness $\mathbf{KP}_e$ along the pressing direction (Fig.~\ref{Graphmimic_second}C(\rom{1})). This 
configuration-dependent stiffness directly shapes how output torque responds to fingertip loading. Under a constant 10 N force, 
total joint torque decreases with flexion, reaching a minimum near 0.67 rad 
(Fig.~\ref{Graphmimic_second}C(\rom{2})). This explains the contrasting control in staccato and legato:
higher stiffness in staccato allows the finger to deliver a rapid force burst, quickly overcoming the 
key's static friction and inertia and release the key swiftly, while lower stiffness facilitates the fine modulation 
of small velocity variations during key depression, enabling smoother transitions between notes, yielding a continuous sound.

The joint kinematics during the performance of \textit{Minuet in G major} (movie S4) further reflect the guidance of the Graph-Mimic 
framework, specifically in its ability to shape natural finger trajectories. For each keystroke, three active joint 
angles (here Abd indicates the abduction/adduction joint) are recorded from two frames before press to two frames 
after release, and each trajectory is temporally normalized to 100 units to enable direct comparison across different 
press durations. Under Graph-Mimic, the index finger MCP executes a natural extension-flexion-extension cycle, with 
smaller PIP/DIP excursions (Fig.~\ref{Graphmimic_second}D(\rom{1})), consistent with human pianists. The corresponding postures (Fig.~\ref{Graphmimic_second}D(\rom{2})) 
appear coordinated and fluid. In contrast, without mimicry, the baseline policy often converges to a high-stiffness motion 
space, resulting in abnormal PIP extension during depression (Fig.~\ref{Graphmimic_second}E(\rom{1}), (\rom{2})). This motor pattern is mechanically 
efficient in generating force, but it leads to non-elegant gestures, deviating from natural 
pianistic motion.

During the key-press phase, Graph-Distance regulates torque and equivalent stiffness by encoding phalangeal vectors 
within the AFG (Fig.~\ref{Overallintro}B), effectively guiding the robotic hand's joint articulation.

To validate its advantage, we compared Graph Distance with Joint Angle Error and End-Effector 
Distance using a representative index finger key-press sequence (Fig.~\ref{Graphmimic_second}F)spanning increasing flexion angles
(HP1-6, RP1-6). The similarity matrices show strong diagonal dominance for both 
Joint Angle Error (Fig.~\ref{Graphmimic_second}G(\rom{1})) and Graph Distance (Fig.~\ref{Graphmimic_second}G(\rom{3})), 
but weak correspondence for End-Effector Distance (Fig.~\ref{Graphmimic_second}G(\rom{2})). While Joint Angle Error captures 
detailed joint rotations, Graph Distance achieves comparable descriptive 
ability by embedding the phalangeal vectors. 

End-Effector Distance, as used in prior work \cite{qian2024pianomime}, considers only fingertip positions and does not 
explicitly capture the internal phalangeal structure. As a result, it may lead to less natural joint 
configurations, such as excessive finger curling in certain joints, which can in turn affect 
articulation quality.

Above two results demonstrate that Graph-Mimic reliably captures the essential kinematic features of
the pre-press phase and implicitly modulates the dynamics behavior of the key-press phase.
This mechanistic advantage is further evidenced by a comparative ablation of the thumb-under maneuver 
(detailed in text S2, fig. \ref{fig:S2}). 
In the absence of Graph-Mimic, the controller relies on suboptimal wrist compensation rather than 
coordinated finger articulation, resulting in severely curled postures and missed keystrokes 
that disrupt temporal continuity. In contrast, Graph-Mimic maintains relaxed finger 
configurations that ensure anthropomorphic style and Operational fluency.

\subsubsection*{Piano acoustics model for tuning musical dynamics}\label{MusicalDynamicsmethods}

Musical dynamics refer to the variations in volume at which notes are played, serving as a vital element of 
expressive performance to convey emotional nuance. By integrating a physics-inspired acoustic model with 
score-consistent loudness optimization, our approach enables expressive dynamic 
control in robotic piano performance.

\begin{figure}
    \centering
    \includegraphics[width=0.85\textwidth]{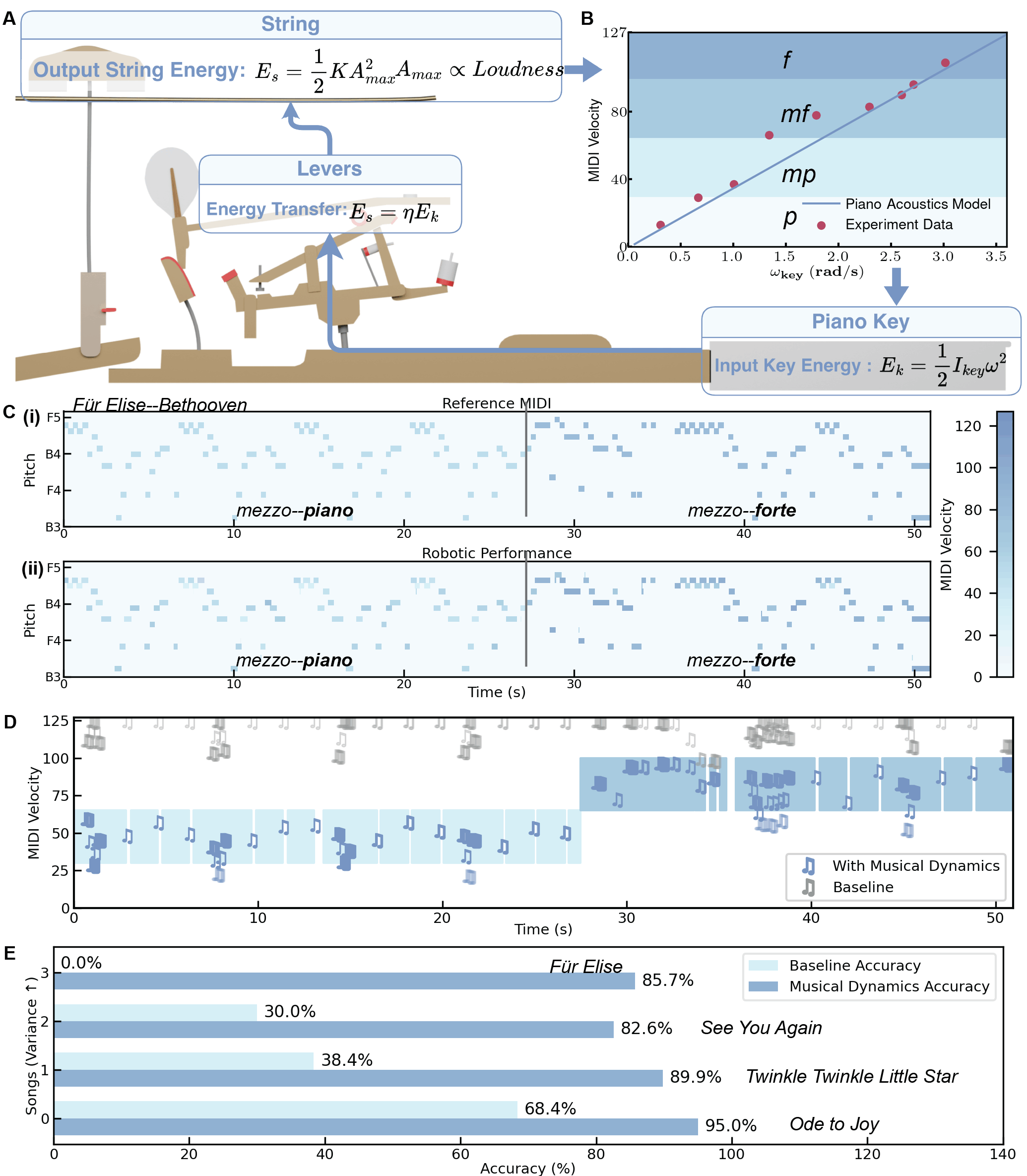}
	\caption{\textbf{Musical dynamics control mechanism and performance.}
    \textbf{(A)} Energy-transfer mechanism of piano sound production, shown as a cross-sectional schematic from key press to hammer motion, initiating string vibration and sound.
    \textbf{(B)} Relationship between the key's angular velocity ($\omega$) and perceived loudness. Four dynamic levels are defined: piano (p), mezzo-piano (mp), mezzo-forte (mf), and forte (f).
    \textbf{(C)} Pitch-loudness heatmap of Beethoven's \textit{Für Elise}: (\rom{1}) reference performance and (\rom{2}) robotic performance.
    \textbf{(D)} Loudness profiles of \textit{Für Elise} performed by the robot, with and without the Musical Dynamics.
    \textbf{(E)} Comparison of four musical pieces ranked by the variance of MIDI velocity, showing velocity accuracy with and without the Musical Dynamics.
    }\label{MusicalDynamics}
\end{figure}

Fig.~\ref{MusicalDynamics}A illustrates the piano's acoustic mechanism: key depression transfers kinetic energy via levers to 
the hammer, which strikes the string to produce vibrations whose amplitude determines loudness.  The loudness of the 
resulting tone is proportional to the maximum amplitude of the string's vibration. 
We adopt MIDI (Musical Instrument Digital Interface) Velocity, ranging from 0 to 127, to quantify this loudness.
Accordingly, a mapping from key angular velocity ($\omega$) to MIDI Velocity(Fig.~\ref{MusicalDynamics}B) shows 
strong agreement between calculated and empirical values (see details in Materials and Methods, text S3, fig. \ref{fig:S3}). 
The curve represents the theoretical relationship based on string dynamics, and measured loudness from a 
piano press experiment (movie S5) confirm the validity of the proposed piano acoustics model.

Building on this model, we categorize the MIDI Velocity range into four standard dynamic levels: 
soft (piano, p), moderately soft (mezzo-piano, mp), moderately loud (mezzo-forte, mf), and loud (forte, f). 
These conventional musical dynamics form the basis for performance evaluation.

Using the simplified piano acoustics model and a loudness-oriented optimization, the robotic hand achieved dynamic loudness control, 
enabling richer emotional expression. For validation, we used \textit{Beethoven's Für Elise}(computer-generated version, movie S6), 
a piece with clear dynamic progressions reflecting emotional intensity. As shown in Fig.~\ref{MusicalDynamics}C(\rom{1}), 
the reference MIDI velocity stays around 49 (mp) in the first half and rises to about 80 (mf) in the second. The trained agent 
reproduces this contrast, with most generated velocities falling within the corresponding dynamic levels 
(Fig.~\ref{MusicalDynamics}C(\rom{2})), confirming that the acoustic mechanism effectively aligns note intensities 
with intended dynamics.

We compared the loudness distributions of robotic performances under our controller and a baseline policy 
(Fig.~\ref{MusicalDynamics}D). The background shading denotes reference loudness levels, with darker intensity indicating 
stronger dynamics. With musical dynamics, most notes align with the reference categories, exhibiting 
minor deviations. In contrast, the baseline policy fails to match any reference levels, striking keys at excessive 
velocity and producing uniformly loud tones devoid of expressive nuance.

To quantify this alignment, we define velocity accuracy as the proportion of notes whose loudness matches the 
reference dynamic category. This metric was evaluated on four pieces with increasing MIDI Velocity 
variance (Fig.~\ref{MusicalDynamics}E). As variance rises, the baseline accuracy drops sharply, reaching zero 
for wide-range pieces. In contrast, our dynamic guidance maintains high consistency, 
achieving 85.7\% accuracy on these challenging works.

\subsubsection*{The piano perceptual test and comparative analysis}
This section presents a piano perceptual test, providing a quantitative way 
to assess expressive qualities that are subjective and difficult to formalize. 

\begin{figure}
    \centering
    \includegraphics[width=1.0\textwidth]{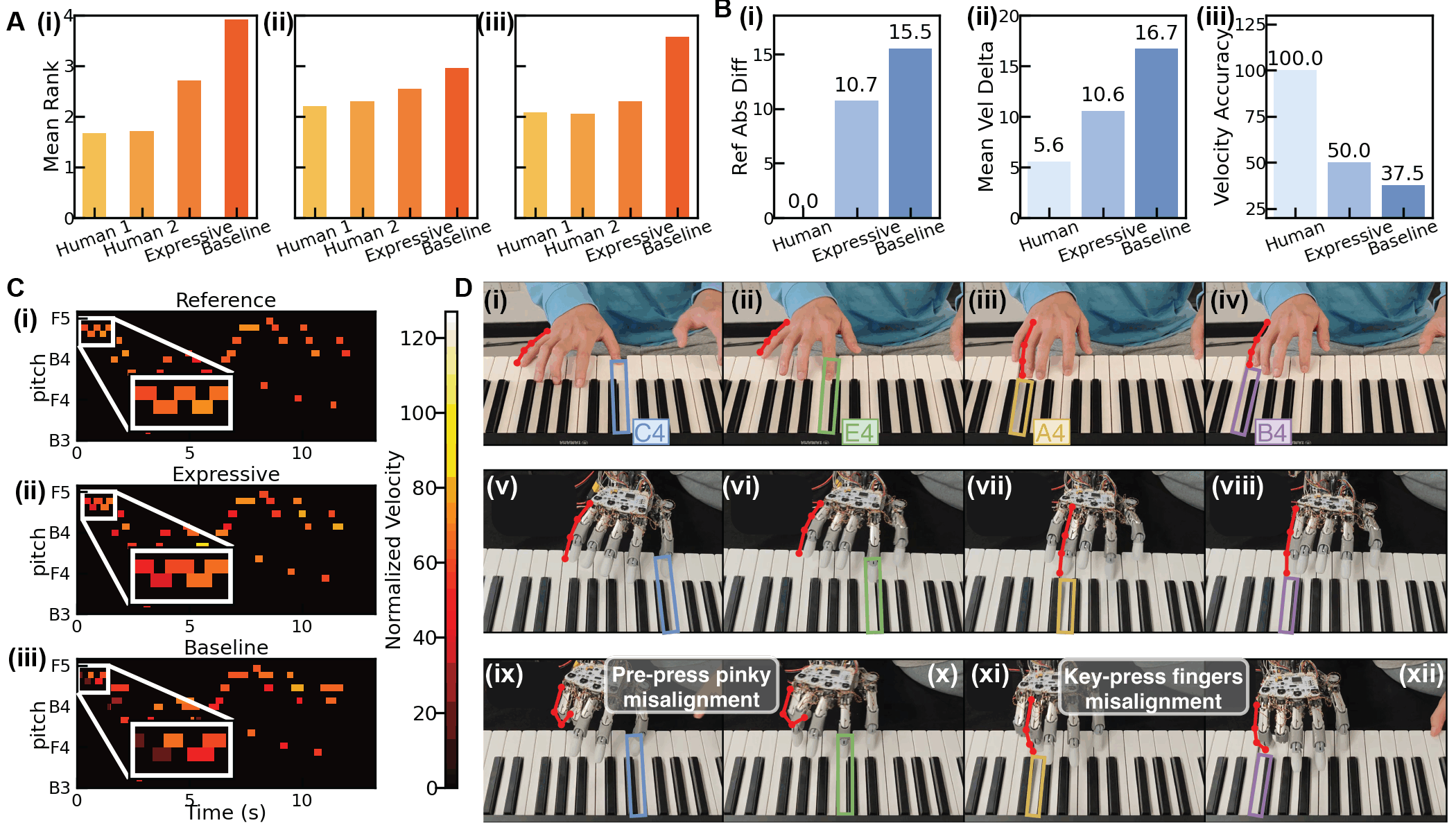}
    \caption{\textbf{Perceptual Test Results and Comparative Analysis of Human, Expressive, and Baseline Performances.}
	\textbf{(A)} The average ranking (1=most preferred, 4=least preferred) 
	across three groups: (\rom{1}) trained pianists, (\rom{2}) general education, and (\rom{3}) untrained individuals. 
	\textbf{(B)} Deviation of loudness from the human reference:(\rom{1}) Reference Absolute Difference, (\rom{2}) Mean Velocity Delta, and (\rom{3}) Velocity Accuracy. 
	\textbf{(C)} Velocity heatmaps for the (\rom{1}) human, (\rom{2}) expressive performance, and (\rom{3}) baseline performance.
	\textbf{(D)} Motion snapshots of the musical passage: (\rom{1})–(\rom{4}) Human reference; (\rom{5})–(\rom{8}) 
	Expressive model; and (\rom{9})–(\rom{12}) Baseline model.}\label{perceptualTest}
\end{figure}

The perceptual test asks participants to rank four audio-only excerpts (two human, expressive robot, baseline robot) 
across three listener groups: formally trained pianists ($n=24$), participants with general piano education ($n=64$), 
and untrained listeners ($n=37$) ((Movie S7, Movie S8)). 
The  excerpt is \textit{Für Elise}, recorded from a human pianist to avoid 
overly rigid playback that would result from a synthesized score.
Average rankings appear in Fig.~\ref{perceptualTest}A(\rom{1})–(\rom{3}), respectively. 
Across all listener groups, the expressive robot performance was ranked ahead of the baseline robot performance, 
although both remained below the human performances.

Overall differences among the four excerpts were evaluated using the Friedman test. Across all listener groups, the Friedman test indicated 
significant differences in rankings (all $p < 0.01$), confirming that participants could reliably distinguish between at least some of the excerpts.
Pairwise comparisons were conducted using Wilcoxon signed-rank tests, with p-values adjusted using the Holm–Bonferroni method to correct for multiple comparisons.
For formally trained pianists, the expressive robot was significantly preferred over the baseline robot ($p < 0.01$) and human 
performances were significantly ranked higher than both robots. General educated participants showed a significant preference difference between Human1 
and the baseline robot ($p=0.0145$). However, comparisons involving the expressive robot and either human performance did not reach statistical significance ($p>0.05$).
For untrained listeners, no statistically significant preference differences were observed between the expressive robot and either human performance ($p > 0.05$). 
In contrast, the baseline robot was significantly less preferred than both human performances ($p < 0.001$), and the expressive 
robot was significantly preferred over the baseline robot ($p < 0.001$). These results indicate that the expressive control substantially 
reduces the perceptual preference gap between robot and human performances.

A comparative analysis of the loudness dynamics and gesture-related expressive behavior 
was conducted to further elucidate the differences between performers.
The Reference Absolute Difference of MIDI Velocity from the human reference is 10.72 for the expressive performance and 
15.50 for the baseline (Fig.~\ref{perceptualTest}B(\rom{1})). Fig.~\ref{perceptualTest}B(\rom{2}) displays the Mean MIDI Velocity Delta 
in MIDI Velocity between consecutive notes, a measure of dynamic stability where larger values indicate greater loudness instability. 
The baseline exhibits higher velocity variability, which correlates with degraded musical expressivity \cite{zhang2011musical}, 
whereas the expressive robot shows only a modest increase (~5 units) over the human reference. Considering that the four primary 
dynamic levels ($p$, $mp$, $mf$, $f$) each span approximately 30 MIDI velocity units, deviations exceeding ~15 units risk altering 
perceived intensity. The baseline surpasses this threshold in both metrics, while the expressive performance remains 
largely within the human dynamic range. Velocity accuracy (Fig.~\ref{perceptualTest}B(\rom{3})) reaches 50\% for the expressive robot 
and 37.5\% for the baseline. Although the absolute accuracy is limited by physical implementation constraints, the expressive 
controller captures essential dynamic characteristics. The clear crescendo at the excerpt opening 
produced by human pianists (Fig.~\ref{perceptualTest}C(\rom{1})) was faithfully reproduced by the expressive robot 
(Fig.~\ref{perceptualTest}C(\rom{2})). This specific loudness shaping feature is absent in the baseline performance 
(Fig.~\ref{perceptualTest}C(\rom{3})), which instead displays unintended soft notes and sudden accents.

Comparing the morphology during pre-press phases, the baseline showed undesirable postures, including excessive curling of the little finger during pre-press 
(Fig.~\ref{perceptualTest}D, \rom{9}–\rom{10}), while the expressive robot produced a more natural extension 
(Fig.~\ref{perceptualTest}D, \rom{5}–\rom{6}) closely matching the human reference 
(Fig.~\ref{perceptualTest}D, \rom{1}–\rom{2}). Morphological differences persisted into key-press: the baseline often left the ring and little fingers overly bent 
(Fig.~\ref{perceptualTest}D, \rom{11}–\rom{12}) following an unnatural straighten–re-bend transition, whereas the expressive robot maintained key-press postures consistent with human behavior 
(Fig.~\ref{perceptualTest}D, \rom{7}–\rom{8}\ vs.\ \rom{3}–\rom{4}), 
transitioning smoothly from extended to flexed states. These human-like gestures and fingering 
strategies also contribute to the visual elegance of the performance.

This analysis is consistent with expert evaluation by a piano professor, who observed 
clear progress in the expressive robotic performance compared to the baseline. In particular, the system demonstrates more 
consistent timing and articulation, improved coordination of finger motions, and a clearer sense of dynamic shaping and 
(text S4).

\subsubsection*{Experimental validation across diverse musical styles}

\begin{figure}
    \centering
    \includegraphics[width=1.0\textwidth]{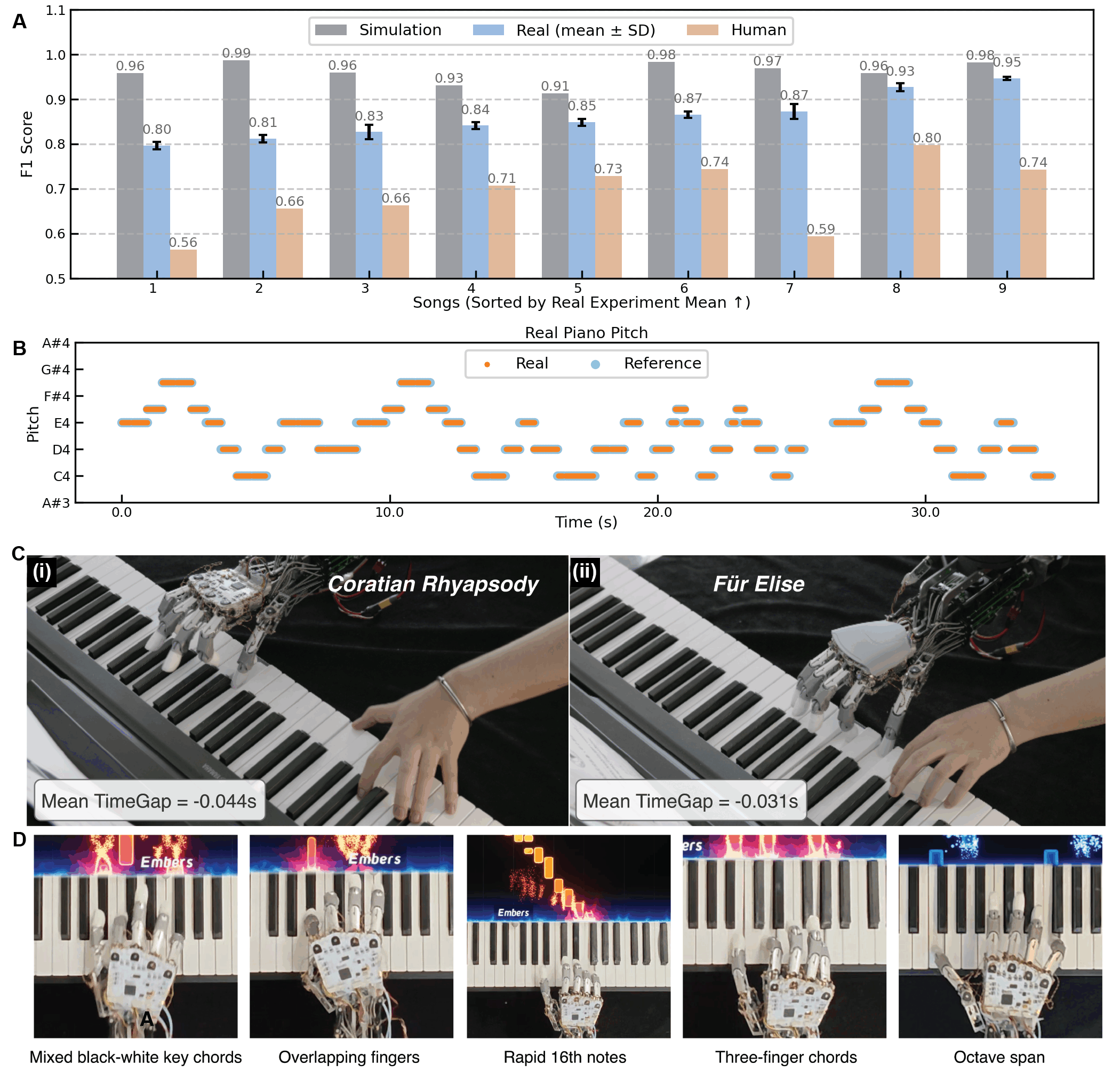}
    \caption{\textbf{Physical performance by the robotic pianist.}
	\textbf{(A)} F1-scores across diverse musical styles in simulation, real-world experiments, and human performances. 
	\textbf{(B)} Pitch-time plots of Ode to Joy, with played notes in orange and reference notes in blue. 
	\textbf{(C)} Duet performances of challenging with human pianists: (\rom{1}) \textit{Croatian Rhapsody} and (\rom{2}) \textit{Für Elise}. 
	\textbf{(D)} Demonstrations of advanced techniques, including mixed black-white key chords, overlapping fingers, rapid sixteenth 
	notes, three-finger chords, and octave spans.}\label{PhysicalResults}
\end{figure}

To assess the efficiency of our framework, we deploy the InReal robotic hand on a 
repertoire spanning classical to popular styles and difficulty levels from Grade 1 to 
7(Fig.~\ref{PhysicalResults}A).
The current deployment is performed in a synchronized sim-and-real execution setting, in which the 
policy received the observation vector in simulation and generated the corresponding control actions, which were 
then executed by the physical robot.

Note-level accuracy is quantified using the F1 score, 
the harmonic mean of precision and recall, where precision and recall are computed 
from true positives (TP, correctly played notes), false positives 
(FP, extra or unintended notes), and false negatives (FN, missed reference notes) (Equation~\ref{eq:precision_recall}). 
\begin{equation}
\begin{aligned}
\text{Precision} &= \frac{TP}{TP + FP},\;
\text{Recall} = \frac{TP}{TP + FN}, \\
\text{F1} &= 2 \times \frac{\text{Precision} \times \text{Recall}}{\text{Precision} + \text{Recall}}
\end{aligned}
\label{eq:precision_recall}
\end{equation}
In simulation, the system achieves near-perfect F1 scores, with an average of 0.96 across 
the evaluated pieces. This is comparable with the 0.94 average reported in \cite{qian2024pianomime}, 
and also above the performance reported in \cite{zhao2024rpm}, where most pieces achieve F1 scores above 0.79. 
Real-world performance remaines highly accurate, with average scores of 0.80-0.95 
across ten trials per piece (Fig.~\ref{PhysicalResults}A, table \ref{tab:piano_scores_transposed}).The high scores 
confirm the effective sim-to-real transfer of the control strategy. This is 
visually corroborated by the pitch-time plot of \textit{Ode to Joy }
(Fig.~\ref{PhysicalResults}B; Movie S3), which exhibits nearly flawless alignment 
with the reference score. Although technically demanding passages 
introduce errors, these did not disrupt the musical continuity, underscoring 
the system's practical reliability. However, human pianists exhibit lower accuracy, 
primarily because it is impossible for them to maintain a perfectly constant, 
unvarying tempo throughout a performance.

To fully demonstrate the piano repertoire, dual performance experiments are conducted where 
a human provided chordal accompaniment for the robot. As shown in Fig.\ref{PhysicalResults}C, 
the renditions of \textit{Croatian Rhapsody} and \textit{Für Elise} achieved musical coherence 
(Movie S9, Movie S10). Temporal synchronization is evaluated 
using the Time Gap metric, defined as the difference between simultaneous note onset times  
(see fig. \ref{fig:S4} and text S5). Given the MIDI acquisition resolution (50 ms), the average Time Gaps 
for \textit{Croatian Rhapsody} and \textit{Für Elise} is -44ms and -31ms, respectively. Both of these 
values fall below the measurement threshold, showing excellent temporal synchronization and 
realism of the human-robot collaboration.

The system reproduce sophisticated techniques essential to human performance 
(Fig.~\ref{PhysicalResults}C). These include mixed black-white key chords 
requiring adaptive hand posture, overlapping finger presses demonstrating 
multi-degree-of-freedom coordination (Movie S11), rapid sixteenth-note passages 
and three-finger chords exhibiting high-speed precision (Movie S12), 
and large octave spans (Movie S13). These examples showcase the system's 
expressive capability, mechanical dexterity, and effective control performance.

\section*{Discussion}

This study addresses expressive robotic piano performance by introducing Graph-Mimic that enables a dexterous 
robotic hand to acquire human-like fingering and touch techniques. In addition, a simplified piano acoustic model 
is integrated to render loudness variations, substantially enhancing the musical expressivity. 
The system successfully executes technically demanding repertoire, 
including Grade 7 works such as Croatian Rhapsody, and demonstrates reliable coordination in duet performance with 
human pianists. On simpler pieces, such as Twinkle, Twinkle, Little Star, the system achieves an F1 score of 0.92, 
substantially outperforming previous approaches that report average F1 scores of approximately 0.60 \cite{zeulner2025learning}. 
Subjective perceptual test among non-expert audiences indicate that the system 
shows no statistical difference between human pianist, while expert evaluations further confirm a significant improvement in expressivity 
compared with baseline controllers. 

Beyond musical performance, this study highlights the broader relevance of piano playing as a benchmark for 
dexterous manipulation. Although piano performance may appear distinct from conventional manipulation tasks, 
it shares three fundamental challenges. First, piano playing involves frequent contact–separation 
interactions between the fingertips and keys, requiring spatiotemporal planning of high degree-of-freedom hands. 
Second, the complex and rapidly varying musical structures, including dense rhythmic patterns such as sixteenth notes, 
impose stringent demands on the response speed of both the robotic hand and its control system. 
Moreover, expressive performance depends on finely regulating keystroke velocity to shape subtle 
dynamic nuances. Together, these characteristics place strict and simultaneous requirements on degrees of freedom, 
speed, and force control, forming a challenging performance triangle for both hardware and algorithms. 
Robotic piano performance provides a structured and quantitative testbed that can substantially 
accelerate the development of dexterous manipulation and advance robots toward real-world deployment.

Finally, this work reveals the advantages of graph-based representations for modeling human hand motion. 
By encoding phalangeal and inter-fingertip vectors, the framework captures the dense interactions inherent in piano performance. A single unified graph 
representation allows the system to learn both pre-press fingering coordination and key-press contact postures.
This graph-based paradigm can extend 
beyond piano performance to general manipulation problems, where incorporating object feature points into 
the graph may enable more natural, efficient, and compliant grasping and manipulation strategies.

\subsubsection*{Limitations and future work}

Several limitations remain. First, the current implementation adopts a four-level loudness classification 
(p/mp/mf/f), which cannot fully capture the dynamic continuum (ppp-fff) essential for nuanced expression. 
Incorporating pressure-sensitive control with high-bandwidth tactile transducers and deep learning-based 
haptic interpretation could enable a continuous and more refined control of dynamics. 

Second, the system currently emphasizes finger dexterity but lacks proximal kinematic control of the arm. 
This limitation restricts weight transfer and joint energy flow required in demanding passages such as Liszt's \textit{Transcendental Étude 
No. 4} and Brahms's \textit{Variations on a Theme of Paganini, Op.35}. Future work could integrate full-arm motion capture with hybrid 
impedance-admittance control, thereby enhancing biomechanical coordination and bringing robotic pianism closer to the expressivity 
of skilled human performers.

Third, the current evaluation framework relies on the F1 score to measure pitch–onset accuracy. 
While useful for assessing execution accuracy, this metric is not well suited to evaluating musical expressivity. 
Human performances naturally contain expressive timing variations that reduce F1 scores.  
In this work, we partially address this limitation by complementing F1 with additional dynamics-related 
metrics, such as Reference Absolute Difference and Mean Velocity Delta, and the perceptual evaluation. 
Nevertheless, a more holistic assessment of expressive piano performance remains an open problem. Future research 
may explore richer evaluation metrics that jointly capture note accuracy, timing flexibility, and dynamic shaping, 
all of which contribute to perceptual expressivity.

Fourth, the current controller still relies on piano-state variables available in simulation, such as key angular velocities. 
Therefore, the present deployment uses simulation observations. Future work could address this limitation either by reconstructing key angular 
velocities from real-time MIDI signals using the simplified piano model, or by distilling a deployable student policy that does 
not require such privileged information from a teacher trained with full simulation states.

Finally, the present work focuses on song-specific policies. 
The insights and control strategies developed through this approach provide a strong foundation for the development 
of generalist piano-playing controllers in future work, potentially enabling real-time and improvisational performance. 

\section*{Materials and Methods}

\subsection*{Study Design}
The primary objective of this study was to evaluate the performance capabilities of a robotic hand 
governed by an expressive controller, benchmarking it against both professional human pianists and a 
baseline robotic controller. The experimental framework comprises two core dimensions. 

A perceptual test involved $n=125$ participants stratified by expertise into formally trained 
pianists ($n=24$), individuals with general piano education ($n=64$), and untrained listeners 
($n=37$), who ranked randomized audio excerpts to determine perceived musical quality. 

Complementing this, an objective technical validation was conducted to assess mechanical 
precision using the F1 score as the primary metric. For physical robot experiments, data 
were aggregated from 10 independent trials per musical piece to account for mechanical variance, 
while reference datasets for both human performances and simulations were derived 
from the highest-scoring instances across multiple recordings to ensure a rigorous and 
competitive benchmark.

\subsection*{Reinforcement Learning Setup}

Reinforcement learning setup is adapted from  RoboPianist project \cite{robopianist2023}. 
Three fundamental reward functions are recommended: Key Press, Energy 
Penalty, and Fingering. The Key Press reward encourages the policy to press and only press the correct keys. Energy 
Penalty discourages high energy expenditure, while Fingering promotes the movement of active fingers as close as 
possible to their corresponding target keys. Full details of the reinforcement learning algorithm and network configuration, including 
architecture and hyperparameters, are provided in  text S6 and table \ref{tab:s4_hyperparameters}.

Building upon these settings, we incorporate two style-oriented reward functions: Graph Distance and Musical 
Dynamics. The full observation, action, and reward setup are summarized in Table~\ref{RLSetUp}.

\subsubsection*{Validation of Graph Distance}

We introduce Graph Distance (GD) metric for quantitative evaluation of morphological 
similarity between robotic and human hand. As depicted in Fig.~\ref{Overallintro}B, human hand 
postures are extracted from piano performance videos on YouTube using MediaPipe\cite{lugaresi2019mediapipe}, 
a real-time machine learning framework capable of identifying 21 anatomical landmarks per hand.  
These landmarks, denoted as $P_{ref} \in  \mathbb{R}^{3 \times n}$ represents the spatial coordinates that capture key 
points along the palm, knuckles, and fingertips, where $n$ is the number of landmarks. From these 
landmarks, 20 feature vectors are derived: 20 vectors represent individual phalanges (four per finger), 
while 4 additional vectors capture 
inter-fingertip spatial relationships. 
These vectors are denoted as $E_{ref}:(e_{ref}^1,\dots,e_{ref}^m ),e_{ref}^i \in \mathbb{R}^{3}$ , where $m$ is 
the total number of vectors. This representation preserves both the kinematic chain of individual 
fingers and the global contour of the hand. Corresponding 
set of 21 feature points on the robotic hand are defined using site elements within the MuJoCo simulator, denoted 
as  $P_{robot} \in  \mathbb{R} ^{3 \times n}$. Feature vectors $E_{robot}:(e_{robot}^1,\dots,e_{robot}^m ), e_{robot}^i \in \mathbb{R} ^{3}$ are 
calculated using the same procedures as with the human hand. To mitigate scale discrepancies between 
human and robotic hands, all vectors are normalized by the distance between the wrist and the tip of 
the middle finger. This yields two comparable Action Frame Graphs: $G_{ref} (P_{ref},E_{ref})$ from the 
human pianist video and $G_{robot} (P_{robot},E_{robot})$ for the robotic hand. The GD is then computed as 
the sum of L2 norms between corresponding AFG vectors:

\begin{equation}
d(G_{\text{ref}}, G_{\text{robot}}) = 
\sum_{i=1}^{m} \sqrt{
\sum_{k=1}^{3} 
\left( e_{\text{ref}_k}^{i} - e_{\text{robot}_k}^{i} \right)^{2}
}
\label{eq:graph_distance_define}
\end{equation}

A notable challenge is that depth inferred from video is typically underestimated. 
This bias is compensated by the Fingering reward, which encourages fingertip motion toward the key 
surface and promotes sufficient key-press depth.
Additionally, The directionality of the fingertip vector is critical for reproducing human-like 
piano performance, as it directly determines the key-press location and resultant tonal quality. However, 
fingertip vectors are also the shortest among phalangeal segments, leading to their influence being 
underrepresented in distance-based similarity metrics. Prior research on human motion perception \cite{basset2022impact} 
highlights that proximal spatial relationships (especially self-contact awareness) shape perceived 
pose equivalence, implying that shorter spatial connections between body parts possess greater 
perceptual salience. In addition to this perceptual motivation, the weighting scheme is designed to account for 
anatomical differences in finger segment lengths. Specifically, we assign weights of 1, 2, 2, and 5 to the metacarpal, proximal, middle, and distal 
phalangeal segments, respectively, following an approximate inverse relationship with segment length.
This design can be interpreted as a normalization of segment-wise contributions in the graph distance metric, 
ensuring that shorter but functionally critical segments are not overshadowed by longer segments.

The final Graph Distance reward is defined as
\begin{equation}
r_{\text{GD}} = g\big(d(G_{\text{ref}}, G_{\text{robot}})\big),
\end{equation}
where \(g(x)\) is a tolerance function:
\begin{equation}
g(x) =
\begin{cases}
1, & l \le x \le u,\\[1.5ex]
\exp\Big[-\frac{1}{2} \big(\frac{d(x)}{m}\, s \big)^2 \Big], & \text{otherwise},
\end{cases}
\end{equation}
with
\begin{equation}
d(x) =
\begin{cases}
l - x, & x < l,\\[1mm]
x - u, & x > u,\\[1mm]
0, & l \le x \le u,
\end{cases}
\end{equation}
and
\begin{equation}
s = \sqrt{-2 \ln(v)}.
\end{equation}

Here, \(l\) and \(u\) denote the lower and upper bounds of the tolerance interval (\([0,5]\) in our 
implementation), \(m\) is the margin controlling the decay width (set to 10), and \(v\) specifies the 
reward value at the margin (set to 0.1). This formulation ensures a smooth, differentiable decay of 
reward as the graph distance exceeds the tolerance range.

\subsubsection*{Piano Acoustics Model and Musical Dynamics Reward}

When a finger strikes a piano key, a complex series of lever mechanisms transmits the motion to a hammer, 
which subsequently strikes the strings, inducing vibrations that produce sound. The loudness of the resulting 
tone is proportional to the vibration amplitude. Energy transfer is a fundamental aspect governing the 
mechanism of piano sound generation. The initial key strike can be approximated as a pivoting motion, with 
the kinetic energy of the key expressed as:
\begin{equation}
E_k = \frac{1}{2}I \omega^2,
\end{equation}
where $I$ is the moment of inertia of the key and $\omega$ is the angular velocity of the key. Empirical data \cite{fletcher2012physics} allow us 
to fit the energy transfer efficiency $\eta $:
\begin{equation}
\eta (\omega) = -0.133\omega ^{-1.009}+0.193,
\end{equation}
which describes how effectively the initial key energy is converted into hammer motion (detailed in text S3, fig. \ref{fig:S3}, table \ref{tab:s2_ke_sound}, table \ref{tab:s3_efficiency}). 
And the effective energy converted into string vibration is $\eta \cdot E_k$, which, at maximum amplitude, can be expressed as:
\begin{equation}
E_{string} = \frac{1}{2} k A_{max}^2,
\end{equation}
where $k$ is the equivalent stiffness for transverse string vibrations and $A_{max}$ is the maximum vibrational amplitude. 
The transverse vibration of a string with fixed ends is governed 
by the following equation:
\begin{equation}
\frac{\partial^2 A}{\partial t^2} = \frac{F}{\mu} \frac{\partial^2 A}{\partial x^2}
\end{equation}
where $A(x,t)$ is the transverse deflection of the string, which is a function of time $t$ and position $x$ along the length of the string. 
$F$ is the tension on the string, and $\mu $ is the linear density. The natural frequency  $\varpi $ of the vibrating system is related to the equivalent 
stiffness $k$ and equivalent mass $m$ by:
\begin{equation}
\varpi = \sqrt{\frac{k}{m}}.
\end{equation}

Considering only the first-order mode of vibration, the equivalent mass is half of the total mass of the string$ m=\mu l/2$, and the 
first-order natural frequency is:
\begin{equation}
\varpi = \frac{\pi}{l} \sqrt{\frac{F}{\mu}}.
\end{equation}
from this, the equivalent stiffness $k$ can be derived as:
\begin{equation}
k = \frac{\pi^2F}{2l}.
\end{equation}
thus, the relationship between sound loudness, quantified as MIDI Velocity, and the key's angular velocity $\omega$ is:
\begin{equation}
\begin{split}
\text{MIDI Velocity} &= p\cdot A_{\max}^2 
= p \cdot \frac{2 \, \eta(\omega) I \, \omega^2 l}{\pi^2 F} \\
&= p \cdot \frac{2 \, \left[-0.133 \, \omega^{-1.009} + 0.193\right] I \, \omega^2 l}{\pi^2 F}
\end{split}
\end{equation}
where MIDI Velocity representing sound loudness, ranging from 0 to 127. The symbol $p$ denotes a scaling constant mapping the 
angular velocity of the keys to the MIDI Velocity value.

To incorporate this model into RL-based piano performance control, the observation space is augmented with angular velocity 
data for all 88 keys. Additionally, the reward function should be refined to 
explicitly account for the deviation between the 
computed MIDI Velocity and the reference MIDI Velocity (\(\text{MIDI Velocity}_{\text{ref}}\)) from the MIDI file:

\begin{equation}
r_{\text{velocity}} = \frac{1}{K} \sum_{i=1}^{K} g\Big( \lVert \text{MIDI Velocity}_{\text{ref}} - \text{MIDI Velocity} \rVert \Big),
\end{equation}

where \(g(x)\) is a tolerance function:
\begin{equation}
g(x) =
\begin{cases}
1, & l \le x \le u,\\[1.5ex]
\exp\Big[-\frac{1}{2} \big(\frac{d(x)}{m}\, s \big)^2 \Big], & \text{otherwise},
\end{cases}
\end{equation}
with
\begin{equation}
d(x) =
\begin{cases}
l - x, & x < l,\\[1mm]
x - u, & x > u,\\[1mm]
0, & l \le x \le u,
\end{cases}
\end{equation}
and
\begin{equation}
s = \sqrt{-2 \ln(v)}.
\end{equation}

Here, \(l\) and \(u\) denote the lower and upper bounds of the tolerance interval (\([0,3]\) in our 
implementation), \(m\) is the margin controlling the decay width (set to 9), and \(v\) specifies the 
reward value at the margin (set to 0.1). This formulation ensures a smooth, differentiable decay of 
reward as the velocity difference exceeds the tolerance range.  

\subsection*{Physical World Setup}

As presented in Fig.~\ref{PhysicalSetup}A, the experimental platform comprises three core components: 
a digital piano, a robotic arm, and a dexterous hand.

\subsubsection*{Piano}
We employ a Yamaha P-48B digital piano, equipped with 88 weighted hammer-action keys that provide a realistic tactile response similar to that of a grand 
piano and produce sound loudness closely approximating a real piano. The instrument 
features stereo-sampled tones with expressive dynamic range. MIDI is transmitted via the USB TO HOST interface, enabling real-time data acquisition 
and quantitative analysis of performance.

\subsubsection*{Robot Arm}
The robotic arm used in this study is the UR5, a 6-degree-of-freedom manipulator with a working envelope of 850 mm and a maximum payload capacity of 
5 kg. Its typical maximum end-effector speed reaches 1 m/s. While sufficient for most musical passages, it imposes limitations when executing rapid 
octave transitions in virtuoso pieces such as \textit{La Campanella}. Nevertheless, the arm provides a high degree of positional repeatability ($\pm $0.1 mm), 
which meets the precision requirements of piano performance. Real-time control is achieved through the RTDE (Real-Time Data Exchange) interface, 
ensuring low-latency communication between the robot controller and our custom control system.

\subsubsection*{Dexterous Robotic Hand}

\begin{figure}
\centering
\includegraphics[width=1.0\textwidth]{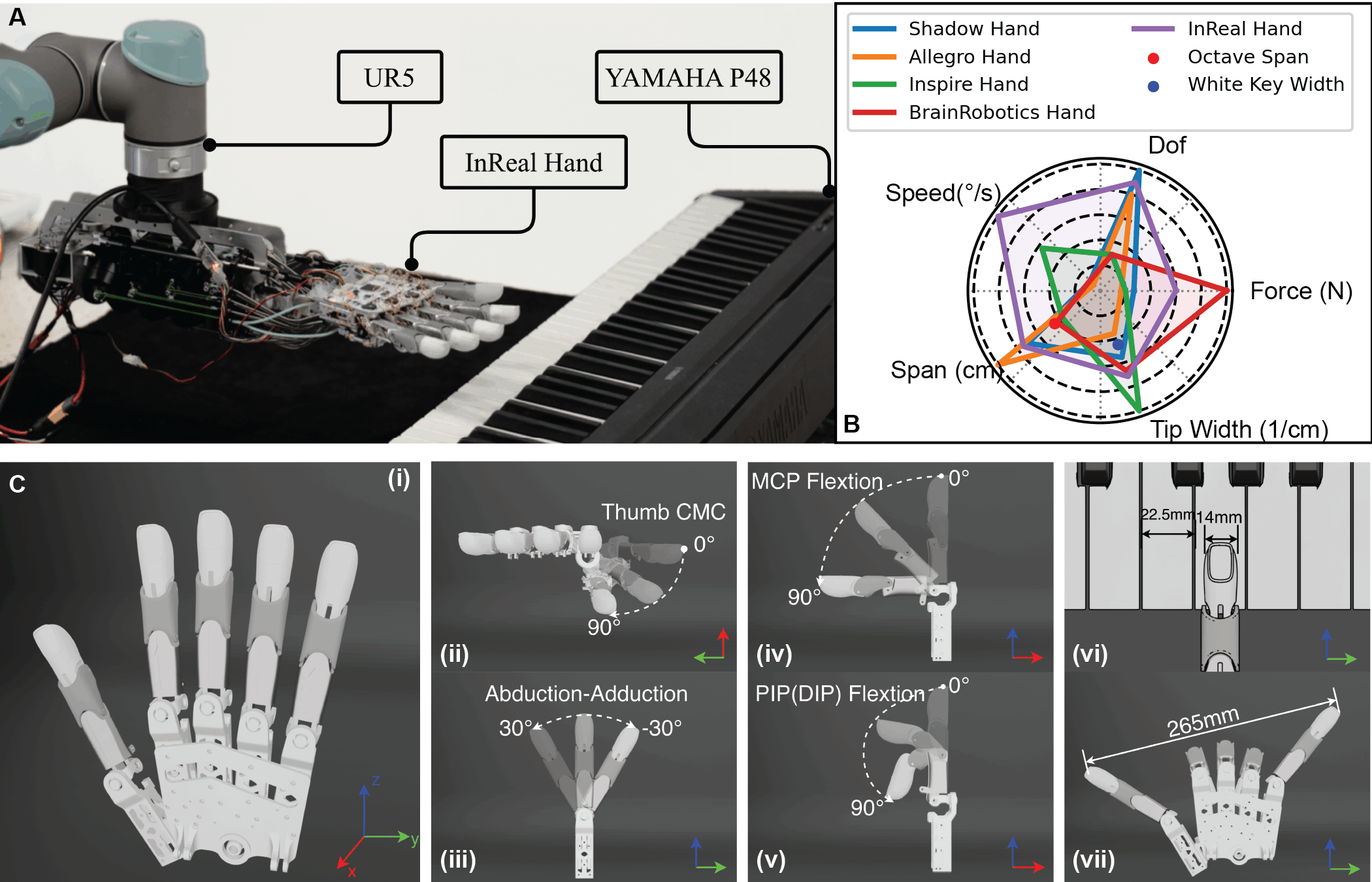}
\caption{\textbf{Experimental setup and InReal robotic hand.}
	\textbf{(A)} Experimental platform comprising a Yamaha P48 digital piano, a UR5 robotic arm, and the InReal robotic hand.
	\textbf{(B)} Comparative evaluation of five robotic hands (Shadow Hand, Bionic Hand, Allegro Hand, Inspire Hand, and InReal Hand) across five key metrics: degrees of freedom, fingertip force, speed, white key width, and octave span.
	\textbf{(C)} Kinematic and structural features of the InReal hand: (\rom{1}) overall view; (\rom{2}) thumb CMC joint mobility ($0^\circ \sim 90^\circ$); (\rom{3}) finger abduction/adduction ($\pm 30^\circ$); (\rom{4}) MCP flexion ($0^\circ \sim 90^\circ$); (\rom{5}) coupled PIP/DIP flexion ($0^\circ \sim 90^\circ$); (\rom{6}) fingertip width (14 mm) relative to white key width (22.5 mm); (\rom{7}) maximum thumb-to-little-finger span (265 mm), exceeding one piano octave ($\approx 157.5$ mm).
	}\label{PhysicalSetup}
\end{figure}

The InReal dexterous robotic hand, weighing 2.1 kg, is well within the payload capacity of the UR5 robotic arm. Its suitability for piano performance 
is systematically evaluated against four other robotic hands \cite{InspireRobots_RH56BFX,BrainCo_DexterousHand,shadowrobot2025,allegrohand2025} 
that have previously been applied to piano-playing tasks (Fig.~\ref{PhysicalSetup}B).

Five dimensions are used for benchmarking:
\begin{enumerate}
    \item \textbf{Degrees of freedom (DoFs):} With 18 active DoFs—including four in the thumb, three in each finger, and two in the wrist (pitch and yaw)—the InReal hand achieves a high degree of dexterity and articulation, closely approximating human biomechanics (Fig.~\ref{PhysicalSetup}C(\rom{2}-\rom{5})).
    \item \textbf{Actuation speed:} The hand exhibits joint speeds exceeding $1000^\circ/s$, enabling a maximum key-strike frequency of $10~Hz$, more than twice the human limit ($\sim 4~Hz$), and supporting rapid passages and trills.
    \item \textbf{Fingertip force:} Powered by external high-torque actuators, the hand delivers peak fingertip forces of up to $18.5~N$, sufficient for expressive dynamics and fortissimo passages without compromising speed.
    \item \textbf{Fingertip width:} Each fingertip measures only $14~mm$, considerably smaller than the $22.5~mm$ width of a white piano key (Fig.~\ref{PhysicalSetup}C(\rom{6})), thereby minimizing accidental key contact and ensuring precision.
    \item \textbf{Maximum span:} The thumb-to-pinky span reaches $265~mm$ (Fig.~\ref{PhysicalSetup}C(\rom{7})), exceeding the $157.5~mm$ octave span of a piano keyboard, and comfortably allowing octave and tenth intervals.
\end{enumerate}

Taken together, the combination of high dexterity, rapid actuation, strong fingertip force, narrow fingertip profile, and large span establishes the 
InReal hand as the most versatile and musically capable robotic hand for piano performance to date.

\subsection*{Statistical Analysis}

Statistical analyses were performed in Python (pandas, SciPy) on participant rank data converted to numeric ranks (1 = most preferred). 
For each listener cohort (formally trained pianists, general-education participants, untrained listeners; 
$n = 24, 64, 37$, respectively), individual rank matrices and group-level mean ranks were computed.

Because each participant provided repeated rankings of four audio excerpts, nonparametric tests for related samples were used. 
Overall differences among the four excerpts were evaluated using the Friedman test, which compares the sum of ranks across participants for each condition. 
The Friedman statistic $\chi_F^2$ is computed as

\begin{equation}
\chi_F^2 = \frac{12}{n k (k+1)} \sum_{j=1}^{k} R_j^2 - 3 n (k+1),
\end{equation}

where
\begin{itemize}
    \item $n$ is the number of participants,
    \item $k$ is the number of conditions (here $k = 4$, corresponding to ``human1'', ``human2'', ``expressive robot'', and ``baseline robot''), 
    \item $R_j$ is the sum of ranks across participants for condition $j$.
\end{itemize}

The p-value of the Friedman statistic was obtained from the $\chi^2$ distribution with $k-1$ degrees of freedom.
Significance was assessed at $\alpha = 0.05$, with $p < 0.05$ indicating that at least two excerpts were ranked significantly differently.

When the Friedman test indicated a significant effect, pairwise comparisons were conducted using the Wilcoxon signed-rank test, which evaluates within-subject differences between two excerpts. 
The Wilcoxon statistic $W$ is calculated as

\begin{equation}
W = \sum_{i=1}^{N} \text{sgn}(d_i) \cdot R_i,
\end{equation}

where
\begin{itemize}
    \item $d_i = x_i - y_i$ is the difference between paired ranks for participant $i$,
    \item $R_i$ is the rank of $|d_i|$ among all non-zero differences,
    \item $\text{sgn}(d_i)$ is the sign of $d_i$ (+1 if $d_i>0$, $-1$ if $d_i<0$),
    \item $N$ is the number of non-zero differences (all participants with tied ranks excluded).
\end{itemize}

Because multiple pairwise comparisons were performed among the four conditions, 
the resulting p-values were adjusted using the Holm–Bonferroni procedure 
to control the family-wise error rate. Corrected p-values ($p$) were reported in \ref{tab:s5_statistic}, 
and statistical significance was defined as $p < 0.05$.  

The participants' preference data is provided in Data S1, 
and the statistical code is available at \url{https://github.com/yanhuhuhahei/Preference-ranking-statistics}.

\subsection*{Supplementary materials}
Supplementary Text S1 to S6\\
Figs. S1 to S5\\
Tables S1 to S6\\
References \textit{(49,50)}\\ 
Movies S1 to S13\\
Legends for movies S1 to S13\\
Data S1



\newpage

\begin{figure}
\centering
\includegraphics[width=0.85\textwidth]{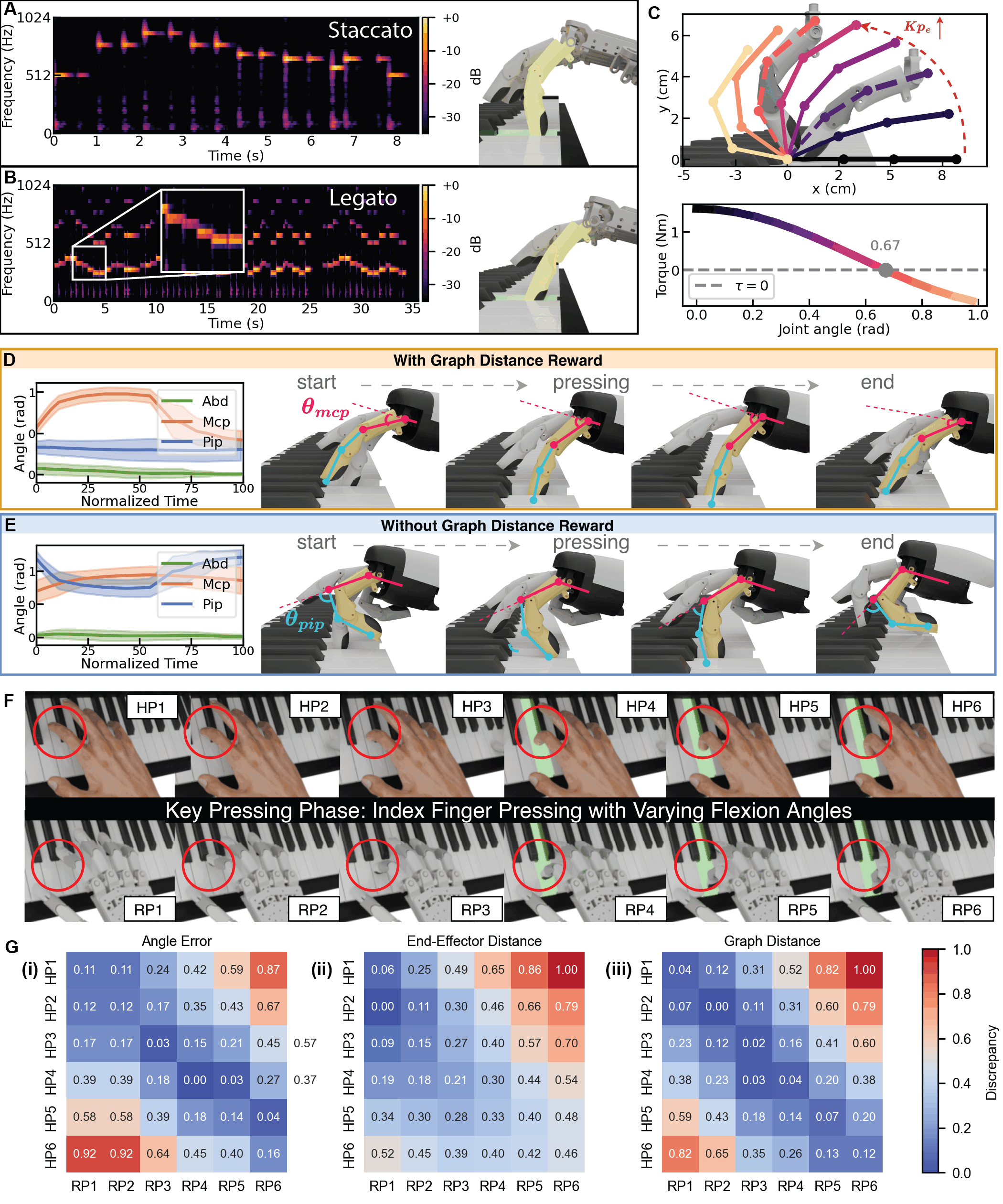}
\caption{\textbf{Graph-Mimic results for key-press performance.}
\textbf{(A)} (\rom{1}) Spectrum of a staccato piece and (\rom{2}) finger postures.
\textbf{(B)} (\rom{1}) Spectrum of a legato piece and (\rom{2}) finger postures.
\textbf{(C)} Linkage-based hand model analysis, detailing (\rom{1}) the Posture-stiffness relationship 
and (\rom{2}) Joint torque versus flexion angle.
\textbf{(D)}–\textbf{(E)}. Index finger key presses with and without Graph Distance guidance: 
(\rom{1}) Joint angle trajectories. (\rom{2}) Sequential frames of finger motion.
\textbf{(F)} Index flexsion snapshots from a human hand (HP1–HP6) and a robotic hand (RP1–RP6).
\textbf{(G)} Normalized discrepancy between human hand and roboitic hand calculated by 
(\rom{1}) joint angle error, (\rom{2}) end-effector distance, and (\rom{3}) graph distance.
}\label{Graphmimic_second}
\end{figure}

\clearpage

\newpage
\begin{table}[h]
\centering
\caption{\textbf{Reinforcement learning setup overview.}\label{RLSetUp}
The observation, action, and reward components of the learning setup, including their meaning and dimensions.}
\label{tab:RLSetup}	
\setlength{\tabcolsep}{8pt} 
\renewcommand{\arraystretch}{1.5} 
\begin{tabular}{p{2.2cm} p{2.5cm} p{8.2cm} p{1.3cm}}
\hline
\textbf{Category} & \textbf{Name} & \textbf{Description} & \textbf{Dim.} \\ 
\hline	
\textbf{Observation} 
 & Piano/key state & Boolean value of piano key activation (1 = active, 0 = inactive). & 88 \\[2pt]
 & Piano/sustain state & Boolean value of sustain pedal activation (1 = active, 0 = inactive). & 1 \\[2pt]
 & Goal & Current and next 10-step piano key and sustain states. & 979 \\[2pt]
 & Fingering & Active hand state for current and next 10 steps. & 55 \\[2pt]
 & Hand Graph & Action frame graph of both human and robot hands. & 120 \\[2pt]
 & Hand joint position & Joint angles of robot hand including wrist translation. & 26 \\[3pt]
\hline	
\textbf{Action} 
 & Hand articulation & Active joints of robotic hand, including wrist translation. & 21 \\[3pt]
\hline	
\textbf{Reward} 
 & Key press reward & $\displaystyle 0.5\!\left(\frac{1}{K}\sum_{i=1}^{K} g(\lVert k_s^i - 1 \rVert_2)\right)
 + 0.5(1 - \mathbf{1}_{\text{false positive}})$ & -- \\[8pt]
 & Fingering reward & $\displaystyle \frac{1}{K}\sum_{i=1}^{K} g(\lVert p_f^i - p_k^i \rVert_2)$ & -- \\[8pt]
 & Energy reward & $\displaystyle |\boldsymbol{\tau}_{\text{joints}}|^{T}|\boldsymbol{v}_{\text{joints}}|$ & -- \\[8pt]
 & Graph distance reward & $\displaystyle \frac{1}{K}\sum_{i=1}^{K} g(\lVert gd_h^i - gd_r^i \rVert_2)$ & -- \\[8pt]
 & Key velocity reward & $\displaystyle \frac{1}{K}\sum_{i=1}^{K} g(\lVert V_{\text{MIDI}}^i - V_r^i \rVert_2)$ & -- \\[8pt]
\hline
\end{tabular}
\end{table}


\clearpage 

%

%
%
%
%
%
%

\noindent
\textbf{Acknowledgments}:
We sincerely acknowledge the open-source project RoboPianist, which laid the foundation 
for the development of this work.We would also like to express our gratitude to Y. Du 
(Zhejiang University) for his participation in the human pianist experiments. 
\textbf{Funding:}
Supported by the STI2030 Major Projects No.\ 2025ZD0218800. 
\textbf{Author contributions:}
Y.L. conceived the core concept, participated in the design and implementation of both 
algorithms and hardware, conducted all experiments, and contributed to manuscript writing. 
X.L. participated in the design and implementation of the robotic hand hardware and was 
responsible for the fabrication of the robotic hand. C.F. and S.C. contributed to the 
development and debugging of the algorithmic framework. Y.Y. was responsible for the 
construction and integration of the experimental platform. C.W. participated in generating human performances.
X.C. provided professional comments on the four tested performances.
Y.J. optimized the conceptual framework and the logical flow of the manuscript. W.Y. and H.W. offered crucial advice 
on the manuscript.
\textbf{Competing interests:}
There are no competing interests to declare.
\textbf{Data and materials availability:}
All data needed to evaluate the conclusions in the paper are present in the 
paper and/or the Supplementary Materials. Specifically, the raw participant 
preference ranking data from the Piano Turing Test are provided in Data S1.




\newpage


\renewcommand{\thefigure}{S\arabic{figure}}
\renewcommand{\thetable}{S\arabic{table}}
\renewcommand{\theequation}{S\arabic{equation}}
\renewcommand{\thepage}{S\arabic{page}}
\setcounter{figure}{0}
\setcounter{table}{0}
\setcounter{equation}{0}
\setcounter{page}{1} 


\begin{center}
\section*{Supplementary Materials for\\ \scititle}

Yanhong Liang$^{1}$,
Xianwei Liu$^{2}$,
Chaojie Fu$^{1}$,
Shaowen Cheng$^{2}$,
Yanyan Yuan$^{1}$,\\
Chengwei Zhuo$^{1}$,
Xi Chen$^{3}$,
Yongbin Jin$^{2\ast}$,
Wei Yang$^{1\ast}$,
Hongtao Wang$^{1\ast}$\\
\small$^{1}$the Center for X-Mechanics, Zhejiang University, Hangzhou \& 310027, China.\\
\small$^{2}$ZJU-Hangzhou Global Scientific and Technological Innovation Center, Hangzhou \& 311200, China.\\
\small$^{3}$Department of Public Physical and Art Education, Zhejiang University, Hangzhou \& 310058, China.\\
\small$^\ast$Corresponding author. Email: yongbinjin@zju.edu.cn\\
\end{center}

\subsubsection*{This PDF file includes:}
Supplementary Text S1 to S6\\
Figures S1 to S5\\
Tables S1 to S6\\
Captions for Movies S1 to S13\\
Captions for Data S1

\subsubsection*{Other Supplementary Materials for this manuscript:}
Movies S1 to S13\\
Data S1

\newpage









\subsection*{Supplementary Text}



\paragraph{Text S1.}
\textbf{Equivalent Stiffness Analysis of the Three-Link Finger Model}

\begin{figure} 
	\centering
	\includegraphics[width=0.6\textwidth]{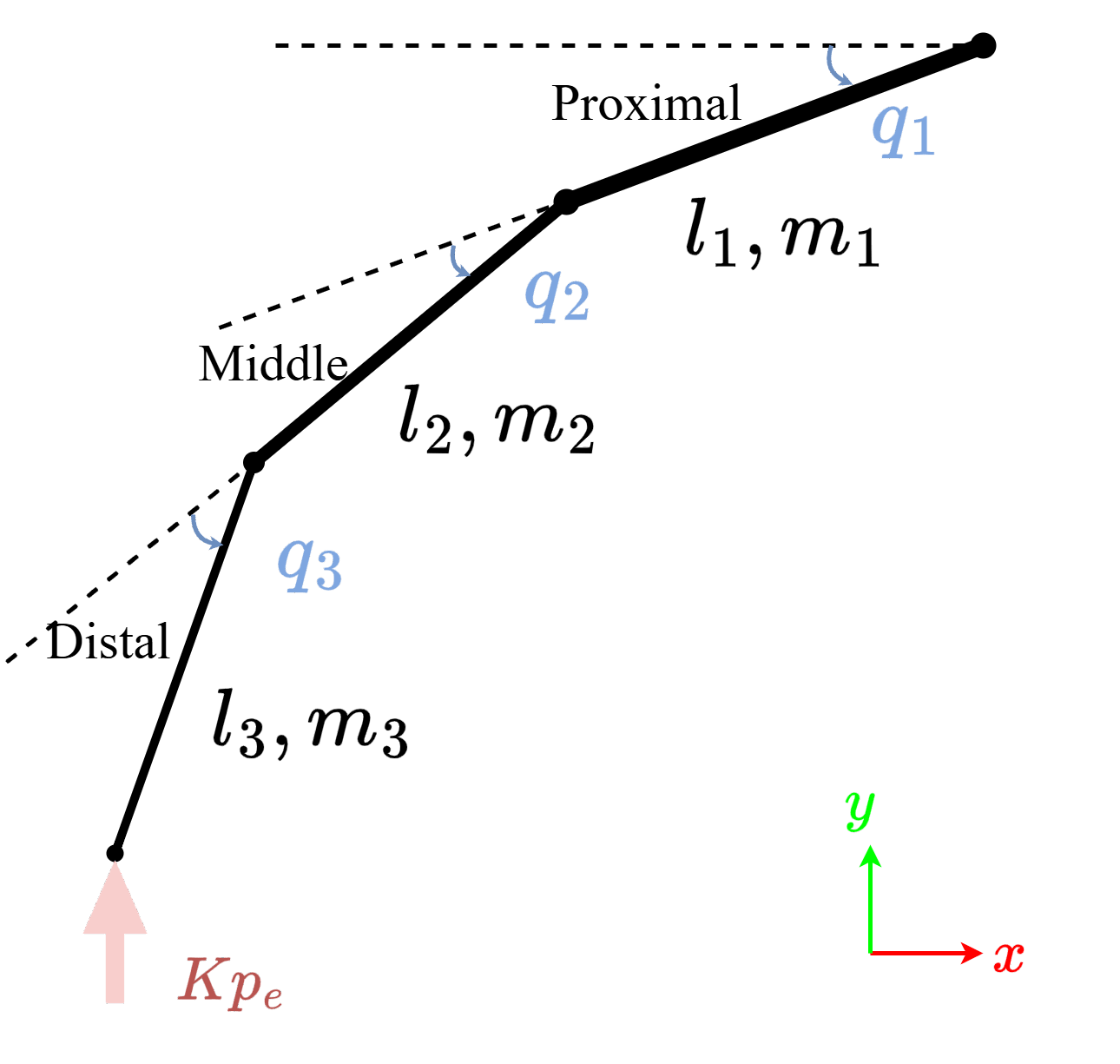} 
	\caption{\textbf{Three-link finger model.}
	The simplified model comprises the metacarpophalangeal (MCP), proximal interphalangeal (PIP), 
	and distal interphalangeal (DIP) joints, 
	with joint angles $\mathbf{q} = [q_1, q_2, q_3]^\mathrm{T}$, 
	link lengths $\mathbf{l} = [l_1, l_2, l_3]^\mathrm{T}$, 
	and link masses $\mathbf{m} = [m_1, m_2, m_3]^\mathrm{T}$. 
	The fingertip stiffness $K_{pe}$ is evaluated along the pressing direction 
	using a Jacobian-based kinematic formulation.}
	\label{fig:S1} 
\end{figure}

To investigate the mechanical origin of posture-dependent stiffness, we approximate the finger as a 
planar three-link chain composed of the metacarpophalangeal (MCP), proximal interphalangeal (PIP), 
and distal interphalangeal (DIP) joints (Fig.\ref{fig:S1}). Each joint is modeled as a revolute joint, 
and the links correspond to the proximal, middle, and distal phalanges. The specific physical 
parameters of these links, including mass, length, and joint stiffness, are detailed in Table \ref{tab:s1_finger_params}.

Let the joint angles be 
\begin{equation}
\mathbf{q} =
\begin{bmatrix}
q_1 \\ q_2 \\ q_3
\end{bmatrix}, 
\quad 
\mathbf{l} =
\begin{bmatrix}
l_1 \\ l_2 \\ l_3
\end{bmatrix}.
\label{eq:joint_and_link}
\end{equation}
	
The fingertip position is expressed as:
\begin{equation}
\mathbf{P}(\mathbf{q}) =
\begin{bmatrix}
x(\mathbf{q}) \\[3pt]
y(\mathbf{q})
\end{bmatrix}
=
\begin{bmatrix}
l_1 \cos q_1 + l_2 \cos(q_1 + q_2) + l_3 \cos(q_1 + q_2 + q_3) \\[3pt]
l_1 \sin q_1 + l_2 \sin(q_1 + q_2) + l_3 \sin(q_1 + q_2 + q_3)
\end{bmatrix}.
\end{equation}

The fingertip Jacobian, mapping joint velocities to fingertip velocities, is given by
\begin{equation}
\mathbf{J}(\mathbf{q}) = \frac{\partial \mathbf{P}}{\partial \mathbf{q}}.
\end{equation}
Each joint is assigned a stiffness matrix
\begin{equation}
\mathbf{K}_q = 
\mathrm{diag}(k_1, k_2, k_3).
\end{equation}
The fingertip stiffness in Cartesian space is obtained as
\begin{equation}
\mathbf{K}_e(\mathbf{q}) = 
\left(\mathbf{J}(\mathbf{q}) \, \mathbf{K}_q^{-1} \, \mathbf{J}(\mathbf{q})^{T}\right)^{-1}.
\end{equation}
This formulation shows that the fingertip stiffness is configuration-dependent:  
as the finger flexes, the Jacobian changes, leading to systematic variation in $\mathbf{K}_e$.

To evaluate the effective stiffness along the vertical pressing direction,  
we project $\mathbf{K}_e$ onto the unit vector
\begin{equation}
\mathbf{d} =
\begin{bmatrix}
0 \\ 1
\end{bmatrix},
\quad
K_e^p(\mathbf{q}) = 
\left( \mathbf{d}^{T} \, \mathbf{K}_e(\mathbf{q})^{-1} \, \mathbf{d} \right)^{-1}.
\end{equation}
This scalar stiffness $K_e^p$ quantifies how resistant the fingertip is to displacement under vertical loading.

The joint torques induced by an external fingertip force $\mathbf{F}$ are obtained from the Jacobian transpose relation:
\begin{equation}
\boldsymbol{\tau}(\mathbf{q}) = 
\mathbf{J}(\mathbf{q})^{T} \, \mathbf{F}.
\end{equation}
Under a constant vertical load of 
\begin{equation}
\mathbf{F} = 
\begin{bmatrix}
0 \\ -10
\end{bmatrix}
\text{ N},
\end{equation}
the total torque magnitude is
\begin{equation}
\tau_{\mathrm{tot}}(\mathbf{q}) = 
\sum_{i=1}^{3} \left| \tau_i(\mathbf{q}) \right|.
\end{equation}

Simulation results (Fig.\ref{Graphmimic_second}) show that $\tau_{\mathrm{tot}}(\mathbf{q})$ decreases as joint flexion increases,  
reaching a minimum near $q_1 = q_2 = q_3 \approx 0.67~\text{rad}$.  
This indicates that deeper flexion reduces the torque demand, consistent with the observed stiffness–torque tradeoff.

\newpage
\paragraph{Text S2.}
\textbf{Comparative Ablation of the Thumb-Under Maneuver}

\begin{figure} 
	\centering
	\includegraphics[width=0.6\textwidth]{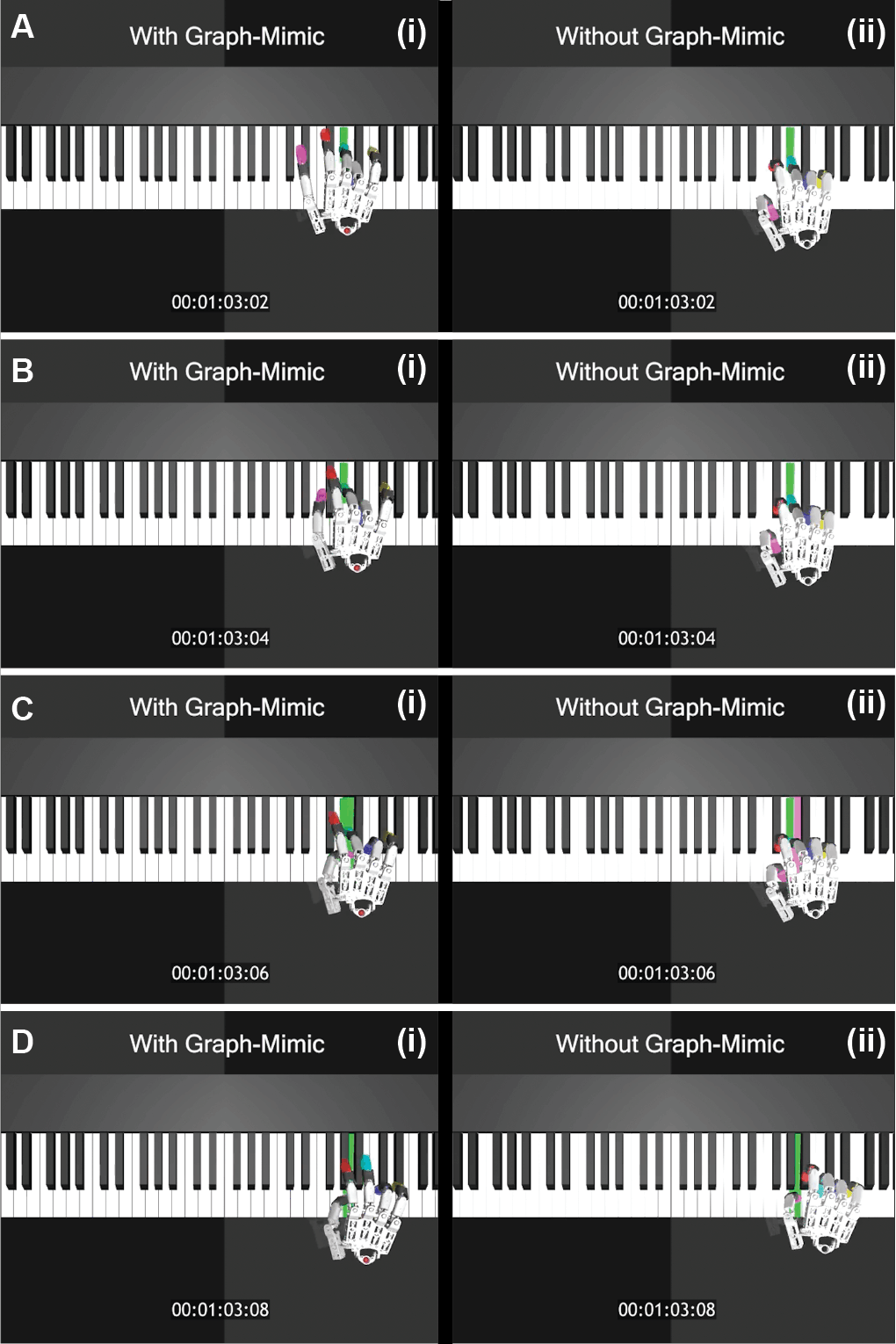} 
	\caption{\textbf{Ablation study of the Graph-Mimic reward during a thumb-under maneuver.}
	 \textbf{(A)} to\textbf{(D)}: Time-lapse sequences (20 ms intervals) comparing robotic 
	 trajectories with (i) and without (ii) the Graph-Mimic framework.}
	\label{fig:S2} 
\end{figure}

The sequential snapshots in Fig.\ref{fig:S2}A–D illustrate a high-difficulty thumb-under transition, 
a maneuver where the thumb crosses beneath the palm to strike a key distal to the middle 
finger’s current position. The Graph-Mimic-enabled controller (Column \rom{1}) maintains 
relaxed and extended finger postures throughout the sequence, prioritizing finger-pad 
contact. In contrast, the baseline controller 
(Column \rom{2}) exhibits excessively curled articulations in both the thumb and index fingers.

Mechanistically, the Graph-Mimic formulation facilitates intra-hand coordination by 
prioritizing finger articulation over wrist translation. Conversely, 
the baseline policy guided solely by fingering rewards tends to seek suboptimal wrist movements 
to satisfy key-finger pair constraints. This reliance on global compensation instead of independent 
finger coordination leads to a missing key at the critical transition point in Fig.\ref{fig:S2}C(\rom{2}). 
Such execution failures disrupt the temporal continuity of the performance, demonstrating 
that Graph-Mimic is essential for resolving the redundant degrees of freedom of the 
anthropomorphic hand into both reliable keystrokes and biologically plausible movement patterns.

\newpage
\paragraph{Text S3.}
\textbf{Energy transfer efficiency in piano action}

Experimental datasets provide complementary perspectives on the energetics of piano key strikes. Table \ref{tab:s2_ke_sound} reports 
the relationship between the kinetic energy imparted to the key and the travel time of the key. Since the vertical displacement 
of the key is essentially fixed by the action geometry, shorter travel times correspond to higher average angular velocities. 
The data reveal that the effective input energy cannot be captured solely by the idealized expression
\begin{equation}
E_k = \tfrac{1}{2} I \omega^2.
\end{equation}
Instead, a more general form is required:
\begin{equation}
E_k \approx \tfrac{1}{2} I \omega^2 + C_1 + C_2 \, T_{\mathrm{travel}},
\end{equation}
where $C_1$ represents the work required to overcome static resistance in the action (independent of speed), 
and $C_2 T_{\mathrm{travel}}$ accounts for velocity-dependent or time-accumulated dissipative losses such as frictional drag. 

For practical modeling, and to maintain computational tractability in simulations, we adopt a simplified form 
by neglecting the dissipative correction terms, as described in the main text:
\begin{equation}
E_k = \tfrac{1}{2} I \omega^2.
\end{equation}
Thus, the experimental measurements in Table \ref{tab:s2_ke_sound} provide the basis for approximating the relationship between 
average angular velocity $\omega$ and input energy.

Table \ref{tab:s3_efficiency} complements this picture by reporting the partitioning of input energy at the key, 
the resulting hammer kinetic energy, and the corresponding transfer efficiency $\eta$ for both grand and upright 
piano actions. In this study we focus on the grand piano data, given their mechanical stability and relevance to 
performance practice. These measurements capture the net efficiency of the action, integrating over inertial, 
frictional, and geometric factors that cannot be fully resolved analytically.

To quantify efficiency as a function of key angular velocity, we combine the datasets: the mapping between 
$\omega(T_{\mathrm{travel}})$ and $E_k$ from Table \ref{tab:s2_ke_sound} establishes the correspondence between strike speed and 
input energy, while Table \ref{tab:s3_efficiency} links input energy to hammer output and transfer efficiency. Fitting the measured 
efficiencies with a power-law regression yields:
\begin{equation}
\eta(\omega) = -0.133 \, \omega^{-1.009} + 0.193,
\end{equation}
which reproduces the nonlinear trend observed in the experiments (Fig.\ref{fig:S2}). The fitted curve provides a compact 
analytic description of how angular velocity modulates energy transfer from key to hammer. This expression is subsequently 
integrated into our acoustic simulations to establish a principled link between key motion and hammer excitation.

\newpage
\paragraph{Text S4.}
\textbf{Expert Commentary on the Four Performances of \textit{Für Elise}}

The evaluation was conducted by Chen Xi, a pianist and Lecturer in the Department of Public Physical 
Education and Art at Zhejiang University. He holds a Doctor of Musical Arts (DMA) in Piano Performance 
and Literature from the Eastman School of Music. His professional distinction includes multiple awards 
in international piano competitions, including First Prize in the Eastern Division of the 2013 MTNA 
(Music Teachers National Association) Competition, and Third Prize and the inaugural prize for 
'Best Francis Poulenc Performance' at the 8th Francis Poulenc International Piano Competition in France. 
Drawing on his combined expertise in performance, analysis, and pedagogy, he provided an expert 
assessment of the selected performances.

\begin{figure} 
	\centering
	\includegraphics[width=1.0\textwidth]{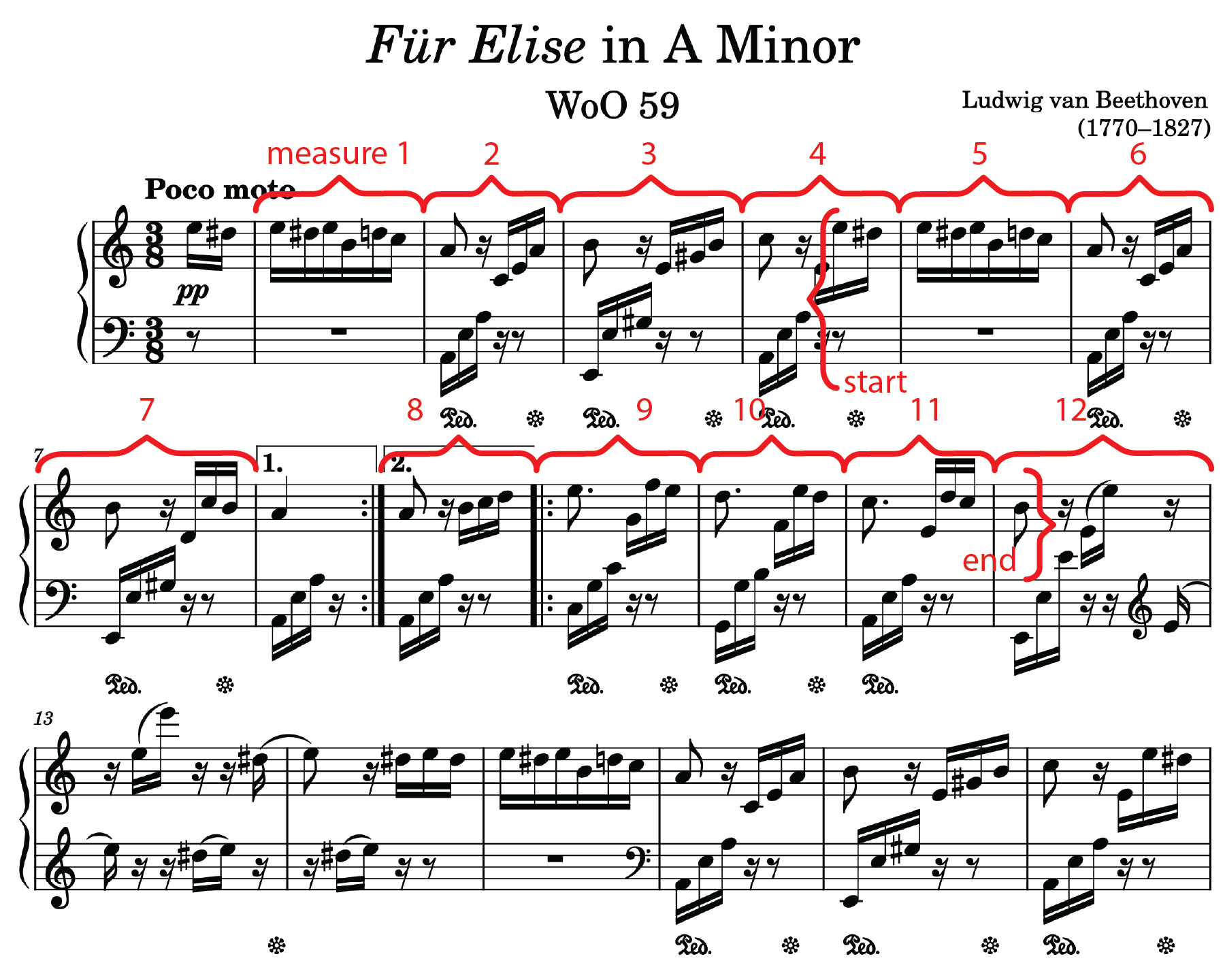} 
	\caption{\textbf{Annotated musical score showing measure segmentation of the evaluated excerpts from \textit{Für Elise}.}
	Annotated musical score of Für Elise indicating the measure boundaries referenced in the expert evaluation. 
	The start and end positions are explicitly marked, and the corresponding measures discussed by the professional 
	piano professor are highlighted for clarity. 
	}
	\label{fig:S5} 
\end{figure}

In evaluating the four performances of Für Elise, attention is directed to right-hand technique 
and expressive delivery, assessed against a high standard of pianistic execution. The following 
observations aim to identify key factors influencing perceptual quality and musical coherence. 
The corresponding excerpt and its measure boundaries are shown in Fig.~\ref{fig:S5}, 
which are consistent with the original score \cite{BeethovenKlavierstuecke_HN12_1986},
available online at \url{https://musescore.com/classicman/fur-elise}.

\paragraph{Performance 1 (Human Reference 1)}
This rendition displays general fluency, though it exhibits moments of inconsistent dynamic control 
and phrasing that deviate from conventional interpretive practices for this style. 
The piece retains its etude-like quality, originally 
composed by Beethoven for pedagogical purposes, with recurring technical patterns intended 
to reinforce specific skills. In measures 2-3, the ascending four-note figure establishes 
a foundational motif repeated throughout. In legato passages such as these, pianists typically 
employ circular wrist motion and weight transfer across fingers to achieve seamless connectivity 
and dynamic shaping. Here, however, the performer relies primarily on vertical finger motion, 
resulting in repeated accents on downbeats and a somewhat abrupt release at phrase endings—most 
noticeably in measure 8, where the final note lacks the graceful taper typical of resolved 
phrasing.

A further technical concern arises in the opening bars, where evenness between the fourth and 
fifth fingers is essential. Optimal execution involves keeping the thumb near the palm to 
stabilize hand balance and minimize tension. In this performance, thumb displacement appears 
to introduce unevenness, particularly audible in the alternating E5 and D$\sharp$5, where the latter 
is accentuated inconsistently.

\paragraph{Performance 2 (Baseline Robotic System)}
This performance demonstrates some phrase shaping but is limited by inconsistencies in timing 
and pitch accuracy. The system struggles to maintain a steady tempo, with occasional note 
omissions and rhythmic instability, especially in measure 5. Nevertheless, several expressive 
details are noteworthy: the B4 in measure 7 is sustained with a curved finger posture—a 
configuration challenging for human players—and is played with higher key-descent velocity, 
followed by a softer, delayed A4 in measure 8. This treatment creates a perceptible resolution 
from supertonic to tonic, reflecting stylistically informed voice leading. Additionally, the 
descending linear motion in measures 9–12 (E5–B4) is rendered with a gradual decrescendo, 
enhancing motif-level articulation and connectivity across repetitions. From an expressive 
standpoint, these dynamics contribute to a more musically coherent rendering than Performance 1, 
despite underlying issues in rhythmic precision.

\paragraph{Performance 3 (Expressive Robotic System)}
This version shows clear progress in reliability and touch control compared to Performance 2. 
Notes are more consistent in timing and articulation, and there is a welcome sense of dynamic 
shaping that gives the playing more musical direction. The hand moves with greater 
coordination—fingers approach and leave the keys with a more natural, fluid motion, avoiding 
the stiff, isolated movements observed in the baseline performance. This is especially 
noticeable in the phrasing around measure 8, where the release of the dominant note into the 
tonic is handled with better continuity, even if the attack itself remains somewhat heavy. 

There is also more discernible dynamic contour, such as the gradual softening in the descending 
line of measures 9–12, which helps outline the phrase. At times, however, the dynamic shifts 
can sound abrupt or overemphasized—such as the strong accent on the C5 at the end of 
measure 5—giving certain moments an exaggerated quality that feels more mechanistic than musical.

Overall, this performance reflects a meaningful step toward more expressive robotic playing. 
The system now responds with better consistency and begins to shape dynamics with intent, 
though the subtler aspects of phrasing and stylistic nuance still call for further refinement 
to achieve a truly natural musical flow.

\paragraph{Performance 4 (Human Reference 2)}
This rendition exhibits fluency and a perceptibly natural tempo flow across phrases, 
contributing to a coherent musical progression throughout the twelve measures. In contrast 
to Performance 1, the pacing between motifs is slightly more organically sustained, supporting 
a stronger sense of structural continuity.

The linear descending motion across measures 9–12 (E5 to B4) is articulated with consistent 
emphasis, each downbeat note played with slightly greater intensity than the preceding three 
notes within the motif. This is achieved through a firmer finger configuration, which lends 
these notes a discernible accentuation. While such an accentual treatment represents a distinct 
interpretative choice rather than a standard stylistic practice, it serves to unify the 
recurring motivic units and reinforces the underlying metrical framework.

Notably, the resolution in measure 8 is executed with less pronounced accentuation compared to 
Performance 1, allowing the phrase to conclude with greater naturalness. Overall, this 
performance demonstrates effective control of tempo and motivic integration, highlighting 
the role of consistent articulatory and dynamic shaping in conveying musical structure.

\newpage
\paragraph{Text S5.}
\textbf{Temporal synchronization metric}

To evaluate temporal coordination in human-robot ensemble performance, we defined a time gap (TG) metric 
(Fig.~S3A). For each note that required simultaneous key depression by the human (chord accompaniment) 
and the robot (melodic line), the TG was calculated as the difference between the robot's onset time and 
the human's onset time, as specified by the reference MIDI score:
\begin{equation}
\mathrm{TimeGap} = t_{\mathrm{robot}} - t_{\mathrm{human}}.
\end{equation}
A positive TG indicates that the human plays after the robot, whereas a negative TG reflects the opposite 
ordering. This differential provides a direct quantitative measure of synchronization precision during 
collaborative performance.

We applied this analysis to two representative pieces: \textit{Croatian Rhapsody} and \textit{Für~Elise}. 
In these tasks, the robotic hand performed the melodic voice while the human played the accompanying 
chords. The piano-roll representations of both ensembles are shown in Fig.~S3B and Fig.~S3C. In the 
illustrations, robot-played notes are marked in blue, human-played chords in red, and notes requiring 
simultaneous execution are highlighted in yellow. This visualization enables clear identification of 
coordination points where TGs are computed, thereby linking the temporal synchronization metric to 
specific musical events.

\newpage
\paragraph{Text S6.}
\textbf{Reinforcement learning algorithm and network configuration}

Our reinforcement learning framework is built upon the open-source infrastructure provided by the 
RoboPianist project \cite{robopianist2023}. Following their established baseline, we employed 
the Soft Actor-Critic (SAC) algorithm \cite{haarnoja2018soft}
, an off-policy actor-critic method based on the maximum entropy reinforcement learning framework. 
This approach aims to maximize the expected cumulative reward while simultaneously maximizing the entropy 
of the policy, ensuring a robust balance between exploration and exploitation in the high-dimensional 
state space of piano performance.

The objective function is defined as:
\begin{equation}
    J(\pi) = \sum_{t=0}^{T} \mathbb{E}_{(s_t, a_t) \sim \rho_\pi} [r(s_t, a_t) + \alpha \mathcal{H}(\pi(\cdot|s_t))]
\end{equation}
where $\alpha$ is the temperature parameter that determines the relative importance of the entropy 
term $\mathcal{H}$ against the reward. Consistent with the RoboPianist implementation, we utilized an automatic 
entropy adjustment mechanism, where $\alpha$ is dynamically tuned during training to maintain a target entropy 
value (set to $-0.5 \times |\mathcal{A}|$, where $|\mathcal{A}|$ is the action dimension).

The implementation utilizes JAX and Flax for high-performance numerical computation. The architecture 
consists of two primary networks: the Actor (Policy) and the Critic (Q-function), along with a trainable 
temperature parameter.

\begin{itemize}
    \item \textbf{Actor Network:} The policy network $\pi_\phi(a|s)$ maps state observations to a probability 
	distribution over actions. It is implemented as a Multi-Layer Perceptron (MLP) with 3 hidden layers of 
	1024 units each. We used GELU activation functions for the hidden layers. The output layer projects to the 
	mean and log-standard-deviation of a diagonal Gaussian distribution. A \texttt{tanh} squashing function is 
	applied to the sampled actions to bound them within the valid action space $[-1, 1]$.
    
    \item \textbf{Critic Network:} The Q-function $Q_\theta(s, a)$ estimates the soft value of state-action 
	pairs. To mitigate Q-value overestimation, we employed the Clipped Double Q-learning technique. The critic 
	comprises an ensemble of 2 independent Q-networks (configured as \texttt{num\_qs=2}). Each Q-network mirrors 
	the actor's structure (3 hidden layers, 1024 units, GELU activation), taking the concatenation of state and 
	action as input.
    
    \item \textbf{Target Networks:} We maintained target networks for the critic, which are updated using Polyak 
	averaging (soft updates) with a coefficient $\tau$ to ensure training stability.
\end{itemize}

To achieve high control accuracy and precise dynamic rendering for the musical piece, we adopted a 
multi-stage training curriculum based on progressive reward shaping: the base policy was first trained 
for $6 \times 10^6$ steps for general execution success, and then subjected to multiple fine-tuning 
rounds (each lasting $3 \times 10^6$ steps) where reward function weights were adjusted to 
prioritize note accuracy and dynamic precision.

The full list of network parameters, optimization settings, and training hyperparameters is detailed 
in Table \ref{tab:s4_hyperparameters}.


\newpage
\subsection*{Supplementary Figures}

\newpage

\begin{figure} 
	\centering
	\includegraphics[width=0.9\textwidth]{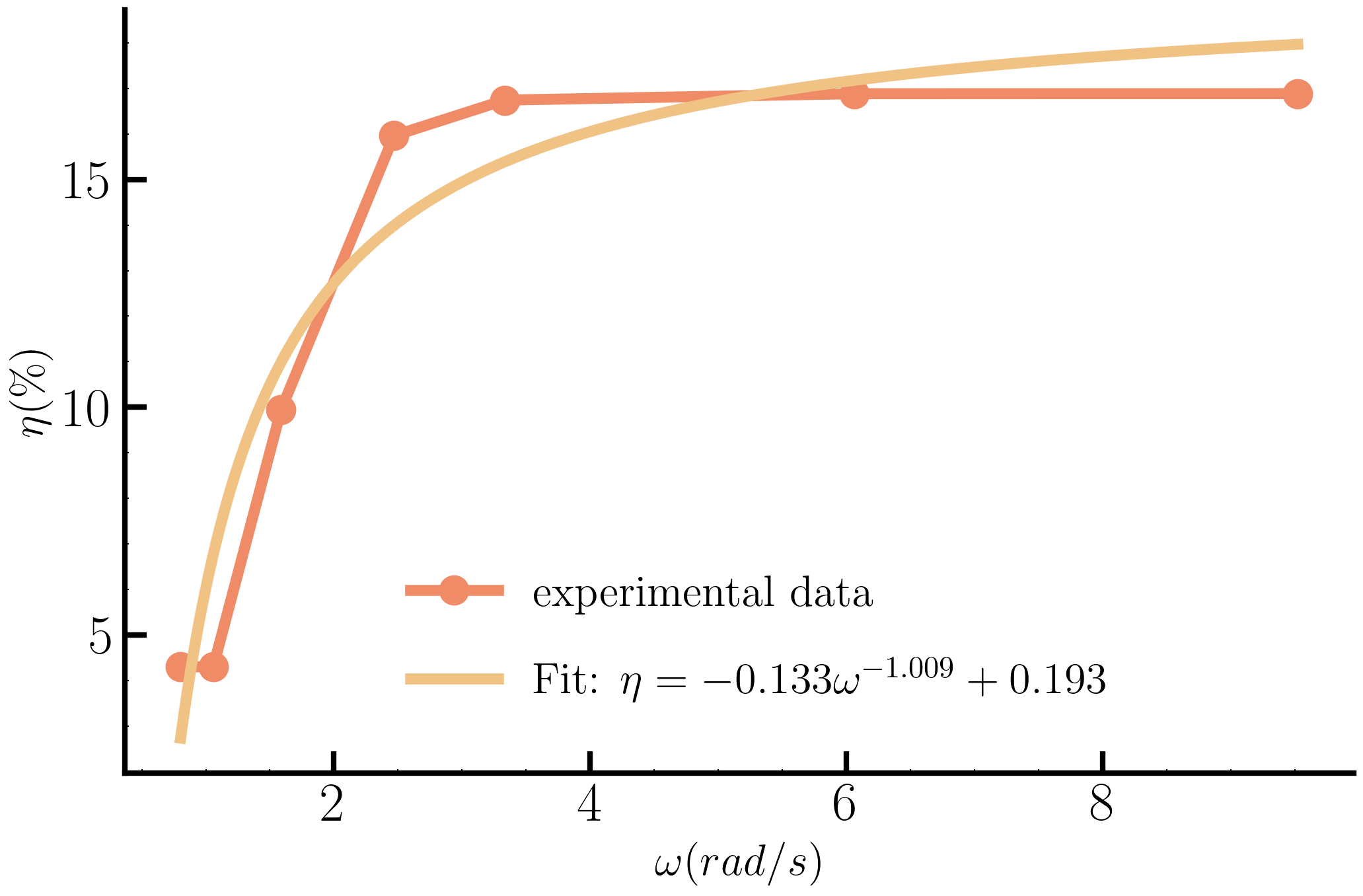} 
	\caption{\textbf{Energy transfer efficiency of piano action.}
	Measured efficiency of energy transfer from key input to hammer motion in a grand piano action (red points 
	and line), plotted as a function of average key angular velocity. The fitted curve (orange line) 
	represents the power-law regression, which captures the nonlinear dependence observed in the 
	experimental data.
	}
	\label{fig:S3} 
\end{figure}

\begin{figure} 
	\centering
	\includegraphics[width=0.9\textwidth]{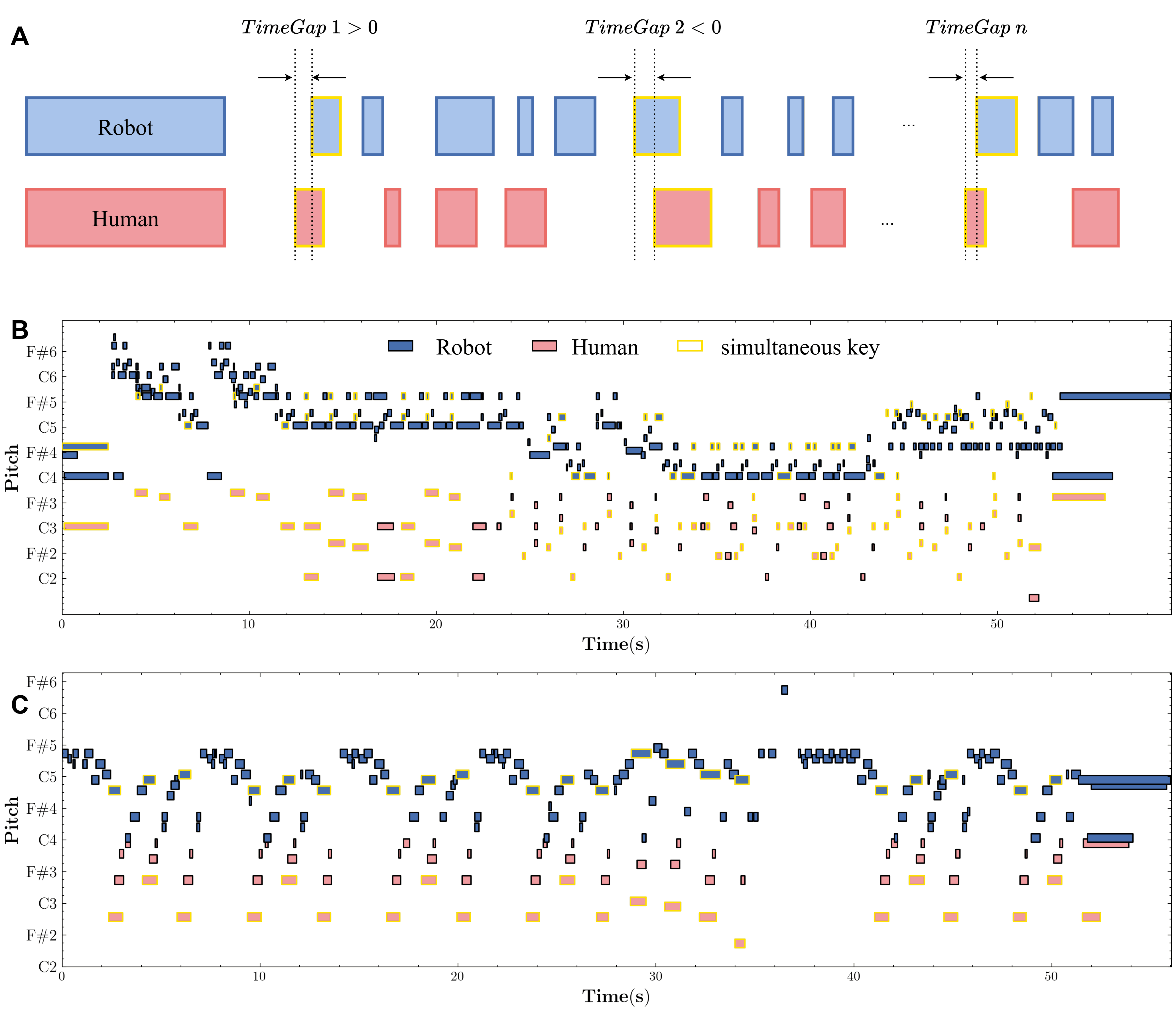} 
	\caption{\textbf{Temporal synchronization in human–robot piano ensemble.}
	(A) Definition of the time gap (TG) metric. Robot-played notes (blue) and human-played notes 
	(red) are aligned against the score; simultaneous notes are highlighted in yellow, and TGs 
	are measured as onset differentials between the two parts. (B) Piano-roll representation of 
	the collaborative performance of Croatian Rhapsody. The robotic hand executed the melodic 
	line (blue), while the human performed chordal accompaniment (red). Simultaneous key events 
	used to compute TGs are marked in yellow. (C) Piano-roll representation of the collaborative 
	performance of Für Elise, with the same color coding as in (B).
	}
	\label{fig:S4} 
\end{figure}

\clearpage
\newpage

\begin{table}[h] 
	\centering
	\caption{\textbf{Finger link parameters used in the three-link model.}
		Each link’s length $l_i$, mass $m_i$, and joint stiffness $k_i$ are listed below.}
	\label{tab:s1_finger_params}

	\begin{tabular}{lccc}
		\\
		\hline
		Link & Length $l_i$ (m) & Mass $m_i$ (kg) & Joint Stiffness $k_i$ (Nm/rad)\\
		\hline
		Proximal & 0.0303 & 0.020 & 10\\
		Middle & 0.0243 & 0.015 & 10\\
		Distal & 0.0272 & 0.010 & 10\\
		\hline
	\end{tabular}
\end{table}

\begin{table}[h] 
	\centering
	\caption{\textbf{Volume of sound related to the kinetic energy given to a piano key.}
		The relationship between key kinetic energy, key travel time, and musical dynamics.}
	\label{tab:s2_ke_sound}

	\begin{tabular}{lcc}
		\\
		\hline
		KE of key (mJ) & Travel time of key (ms) & Musical dynamic\\
		\hline
		7 & 83 & pp\\
		10 & 63 & mp\\
		20 & 42 & mf\\
		50 & 27 & f\\
		100 & 20 & mff\\
		200 & 11 & ff\\
		290 & 7 & fff\\
		\hline
	\end{tabular}
\end{table}

\newpage

\begin{table}[h] 
	\centering
	\caption{\textbf{Mechanical efficiency of grand and upright piano actions.}
		Comparison between kinetic energy transfer from key to hammer in the two piano types.
		KE denotes the kinetic energy of the hammer, and Eff. denotes the mechanical efficiency.}
	\label{tab:s3_efficiency}

	\begin{tabular}{lcccc}
		\\
		\hline
		Input to key (mJ) & \multicolumn{2}{c}{Grand action} & \multicolumn{2}{c}{Upright action}\\
		\cline{2-5}
		& KE (mJ) & Eff. (\%) & KE (mJ) & Eff. (\%)\\
		\hline
		10 & 0.44 & 4.4 & 0.55 & 5.5\\
		20 & 0.87 & 4.3 & 0.91 & 4.5\\
		30 & 3.05 & 10.2 & 2.93 & 9.8\\
		40 & 5.20 & 13.0 & 4.52 & 11.3\\
		50 & 7.31 & 14.6 & 6.13 & 12.3\\
		60 & 9.57 & 15.9 & 7.32 & 12.2\\
		80 & 13.25 & 16.6 & 10.31 & 12.9\\
		100 & 16.73 & 16.7 & 13.65 & 13.6\\
		120 & 20.23 & 16.9 & 16.86 & 14.0\\
		\hline
	\end{tabular}
\end{table}

\begin{table}[h]
    \centering
    \caption{\textbf{Performance Scores for Each Track across Human, Simulation, and Real-world Trials.}}
    \label{tab:piano_scores_transposed}
    \resizebox{\columnwidth}{!}{
        \begin{tabular}{l|ccccccccc}
            \hline
            \textbf{Trial Type} & \textbf{S1} & \textbf{S2} & \textbf{S3} & \textbf{S4} & \textbf{S5} & \textbf{S6} & \textbf{S7} & \textbf{S8} & \textbf{S9} \\
            \hline
            Human & 0.5643 & 0.6559 & 0.6628 & 0.7073 & 0.7289 & 0.7439 & 0.5940 & 0.7982 & 0.7431 \\
            Simulation & 0.9588 & 0.9871 & 0.9592 & 0.9315 & 0.9129 & 0.9838 & 0.9694 & 0.9578 & 0.9825 \\
            \hline
            Real-1 & 0.8072 & 0.8041 & 0.8104 & 0.8516 & 0.8437 & 0.8591 & 0.8972 & 0.9231 & 0.9466 \\
            Real-2 & 0.7959 & 0.8103 & 0.8427 & 0.8326 & 0.8543 & 0.8680 & 0.8901 & 0.9160 & 0.9481 \\
            Real-3 & 0.7782 & 0.8092 & 0.8167 & 0.8394 & 0.8510 & 0.8598 & 0.8931 & 0.9355 & 0.9418 \\
            Real-4 & 0.8029 & 0.8157 & 0.8567 & 0.8488 & 0.8402 & 0.8714 & 0.8832 & 0.9237 & 0.9498 \\
            Real-5 & 0.8096 & 0.8041 & 0.8397 & 0.8420 & 0.8436 & 0.8640 & 0.8599 & 0.9344 & 0.9437 \\
            Real-6 & 0.7981 & 0.8101 & 0.8243 & 0.8369 & 0.8397 & 0.8733 & 0.8591 & 0.9157 & 0.9447 \\
            Real-7 & 0.7948 & 0.8186 & 0.8088 & 0.8500 & 0.8411 & 0.8756 & 0.8776 & 0.9237 & 0.9463 \\
            Real-8 & 0.7914 & 0.8261 & 0.8314 & 0.8460 & 0.8532 & 0.8715 & 0.8624 & 0.9312 & 0.9415 \\
            Real-9 & 0.7936 & 0.8252 & 0.8347 & 0.8319 & 0.8633 & 0.8553 & 0.8529 & 0.9435 & 0.9542 \\
            Real-10 & 0.7924 & 0.8001 & 0.8046 & 0.8324 & 0.8567 & 0.8581 & 0.8500 & 0.9224 & 0.9460 \\
            \hline
        \end{tabular}
    }

    \vspace{0.8em}
    \begin{flushleft}
        \footnotesize
        \textbf{Note:} Tracks S1--S9 are sorted by the mean score of their 10 real-world trials in ascending order. \\
        \textbf{Track IDs:} 
        \textbf{S1}: Croatian Rhapsody; 
        \textbf{S2}: River Flows in You; 
        \textbf{S3}: The Truth That You Leave; 
        \textbf{S4}: Sonata No. 16, K.545; 
        \textbf{S5}: See You Again; 
        \textbf{S6}: Für Elise; 
        \textbf{S7}: Minuet in G major; 
        \textbf{S8}: Twinkle Twinkle Little Star; 
        \textbf{S9}: Ode to Joy.
    \end{flushleft}
\end{table}

\begin{table}[h]
    \centering
    \caption{\textbf{Hyperparameters for Soft Actor-Critic.}
        The specific network architecture, training configuration, and algorithm parameters used in the experiments.}
    \label{tab:s4_hyperparameters}

    \begin{tabular}{lc} 
        \hline
        Parameter & Value\\ 
        \hline
        \multicolumn{2}{l}{\textit{Network Structure}} \\ 
        Hidden Layers (Actor \& Critic) & $[1024, 1024, 1024]$\\
        Activation Function & GELU\\
        Optimizer & Adam\\
        Optimizer $\beta_1, \beta_2$ & $0.9, 0.999$\\
        \hline
        \multicolumn{2}{l}{\textit{Training Configuration}} \\ 
        Base Policy Training Steps & $6 \times 10^6$\\
        Reward Fine-tuning Steps & $3 \times 10^6$\\
        Replay Buffer Capacity & $10^6$\\
        Minibatch Size & 256\\
        Warmup (Seed) Steps & 5000\\
        Critic Target Update Frequency & 1\\
        Actor Update Frequency & 1\\
        \hline
        \multicolumn{2}{l}{\textit{Learning Rates}} \\ 
        Actor Learning Rate & $3 \times 10^{-4}$\\
        Critic Learning Rate & $3 \times 10^{-4}$\\
        Temperature Learning Rate & $3 \times 10^{-4}$\\
        \hline
        \multicolumn{2}{l}{\textit{Algorithm Parameters}} \\ 
        Discount Factor & 0.8\\
        Soft Update Coefficient & 0.005\\
        Target Entropy & $-0.5 \times \text{dim}(\mathcal{A})$\\
        Initial Temperature & 1.0\\
        Number of Q-networks & 2\\
        Critic Dropout Rate & 0.0\\
        Critic Layer Norm & False\\
        Actor Log Std Bounds & $[-20, 2]$\\
        \hline
    \end{tabular}
\end{table}

\begin{table}[t]
\centering
\caption{\textbf{Preference ranking statistics.}
Friedman tests and Holm-corrected Wilcoxon signed-rank post-hoc comparisons 
for participant preference rankings across three listener cohorts. 
Friedman and Wilcoxon tests report p-values, with statistical significance defined 
as $p < 0.05$.}
\label{tab:s5_statistic}
\begin{tabular}{lllll}
\hline
Group &  $p_{\mathrm{raw}}$ & $p$ & Significant \\
\hline

\multicolumn{4}{l}{\textbf{Formal training (n = 24)}} \\
Friedman (all tracks)  & 0.00000 & -- & True \\
Human1 vs Human2 & 0.876314 & 0.876314 & False \\
Human1 vs Expressive Robot & 0.001990 & 0.005969 & True \\
Human1 vs Baseline Robot & 0.000014 & 0.000080 & True \\
Human2 vs Expressive Robot & 0.003236 & 0.006472 & True \\
Human2 vs Baseline Robot & 0.000013 & 0.000080 & True \\
Expressive vs Baseline Robot & 0.000021 & 0.000083 & True \\
\hline

\multicolumn{4}{l}{\textbf{General education (n = 64)}} \\
Friedman (all tracks)  & 0.00477 & -- & True \\
Human1 vs Human2             & 0.712852  & 0.712852 & False \\
Human1 vs Expressive Robot   & 0.115312  & 0.345936 & False \\
Human1 vs Baseline Robot     & 0.002422  & 0.014533 & True \\
Human2 vs Expressive Robot   & 0.226771  & 0.453542 & False \\
Human2 vs Baseline Robot     & 0.017156  & 0.085782 & False \\
Expressive vs Baseline Robot & 0.059446  & 0.237786 & False \\
\hline

\multicolumn{4}{l}{\textbf{Untrained individuals (n = 37)}} \\
Friedman (all tracks)  & 0.00000 & -- & True \\
Human1 vs Human2             & 0.859751& 1.000000& False \\
Human1 vs Expressive Robot   & 0.399292& 1.000000& False \\
Human1 vs Baseline Robot     & 0.000028& 0.000139& True \\
Human2 vs Expressive Robot   & 0.369101& 1.000000& False \\
Human2 vs Baseline Robot     & 0.000010& 0.000060& True \\
Expressive vs Baseline Robot & 0.000157& 0.000627& True \\
\hline

\end{tabular}
\end{table}


\clearpage 

\paragraph{Caption for Supplementary Movie S1.}
\textbf{\textit{River Flows in You} pre-press fingering demonstration.}
Highlights five fingering techniques, including consecutive fingering, adaptive span adjustment, thumb-under, 
and finger-over transitions, illustrating human-like hand motion during the pre-press phase.

\paragraph{Caption for Movie S2.}
\textbf{Staccato articulation in \textit{Twinkle Twinkle Little Star}.}
Demonstrates crisp note separation achieved through vertical fingertip contact and rapid release, as guided by the Graph-Mimic controller.

\paragraph{Caption for Movie S3.}
\textbf{Legato articulation in \textit{Ode to Joy}.}
Shows smooth note transitions enabled by overlapping key depressions and stable finger-pad contact under Graph-Mimic guidance.

\paragraph{Caption for Movie S4.}
\textbf{Finger joint kinematics during robotic performance of \textit{Minuet in G major}.}
Compares finger–key contact joint trajectories under the proposed Graph-Mimic controller and the baseline policy, 
highlighting differences in coordination and motion naturalness.

\paragraph{Caption for Movie S5.}
\textbf{Experimental measurement of key velocity–loudness mapping.}
Shows the electronic piano measurements used to validate the relationship between key angular velocity and MIDI Velocity across a wide range of keystroke speeds.

\paragraph{Caption for Movie S6.}
\textbf{Expressive dynamic control in \textit{Für Elise}.}
Shows accurate reproduction of dynamic contrasts through velocity modulation aligned with reference MIDI dynamics.

\paragraph{Caption for Movie S7.}
\textbf{Audio-only excerpts for the Piano Turing Test.}
Four audio clips are presented without visual information, including two performances by a human pianist, 
one by the baseline robotic system, and one by the proposed expressive robotic system. 
Participants were asked to rank the excerpts according to their listening preference.

\paragraph{Caption for Movie S8.}
\textbf{Visualized performances corresponding to the Piano Turing Test.}
The video presents the same four performances as in Movie S7 with visual information: 
(No.1) human pianist, (No.2) baseline robotic performance, (No.3) proposed expressive robotic performance, and (No.4) human pianist.

\paragraph{Caption for Movie S9.}
\textbf{Human–robot ensemble performance of \textit{Croatian Rhapsody}.}
Shows seamless acoustic integration and timing coordination between the human performer and the robotic piano system in a Grade 7 repertoire.

\paragraph{Caption for Movie S10.}
\textbf{Human–robot ensemble performance of \textit{Für Elise}.}
Shows seamless acoustic integration and timing coordination between the human performer and the robotic piano system.

\paragraph{Caption for Movie S11.}
\textbf{Mixed black–white key chords and overlapping finger presses.}
Demonstrates adaptive hand postures during mixed black–white key chord execution and overlapping 
finger presses in \textit{The Truth That You Leave}, requiring coordinated multi degree-of-freedom control.

\paragraph{Caption for Movie S12.}
\textbf{High-speed passages and multi-finger chord execution.}
Shows rapid sixteenth-note passages and three-finger chord performance 
in \textit{Sonata No.16, 1st Movement, K.545}, highlighting high-speed precision and temporal control.

\paragraph{Caption for Movie S13.}
\textbf{Large-span octave execution.}
Illustrates sustained octave spans in \textit{See You Again}, demonstrating extended reach, 
hand reconfiguration, and stability during wide-range key transitions.

\paragraph{Caption for Data S1.}
\textbf{Preference ranking data from the Piano Turing Test.}
This dataset is provided as a compressed archive (.zip) containing three CSV files: 
formally-trained.csv, general-educated.csv, and untrained.csv, corresponding to the three 
listener groups analyzed in the study. Each file contains the timestamp, device operating 
system (Device\_OS), anonymized participant identifier (Participant\_ID), and the preference 
ranking (Preference\_Ranking).To protect participant privacy, the Participant\_ID was generated 
by applying a SHA-256 cryptographic hash function to the original IP addresses, ensuring anonymity 
while allowing for the verification of unique respondents. Note on data collection: For the untrained.csv 
dataset, the last five entries share an identical Participant\_ID; these responses were collected 
offline using a single shared device at a designated testing station.
The Preference\_Ranking column lists the order of preference (from most preferred to least preferred) 
for four anonymized versions of the performance. The decoding key for the stimuli is as follows:
No. 1: Human Performance 1.
No. 2: Baseline Robotic Performance.
No. 3: Expressive Robotic Performance (Proposed method with Graph-Mimic and Musical Dynamics).
No. 4: Human Performance 2.



\end{document}